\documentclass[11pt]{article}

\usepackage[preprint]{acl}

\usepackage{times}
\usepackage{latexsym}

\usepackage[T1]{fontenc}

\usepackage[utf8]{inputenc}

\usepackage{microtype}
\usepackage{twemojis}
\usepackage{inconsolata}

\usepackage{graphicx}

\usepackage{subcaption}
\usepackage{booktabs}
\usepackage{multirow} 
\usepackage{array}    
\usepackage{makecell}
\usepackage{subcaption}
\usepackage{tabularx}
\usepackage{multirow}
\usepackage{adjustbox}
\usepackage{siunitx}
\usepackage{pifont}
\usepackage{amsmath}
\usepackage{amssymb}
\usepackage{titling}
\usepackage{algorithm}
\usepackage{algorithmic}
\title{What Do Hallucinations Reveal About Multimodal Reasoning? Diagnosing Visual Grounding Failures via Contrastive Decoding Probes}

\author{
  \textbf{Zhipeng Zhao}\textsuperscript{1},
  \textbf{Wenxu Wang}\textsuperscript{1},
  \textbf{Peishun Liu}\textsuperscript{1},
  \textbf{Ruichun Tang}\textsuperscript{1,*}
\\
\\
  \textsuperscript{1}Ocean University of China
\\
  \small{\texttt{\{zhaozhipeng\}@stu.ouc.edu.cn}} \\
  \small{\texttt{\{wangwenxu, liups, tangruichun\}@ouc.edu.cn}}
\\
  \small{\textsuperscript{*}Corresponding author}
}

\begin{document}
\maketitle
\begin{abstract}
When strong multimodal models are widely available, progress requires new scientific methodologies beyond benchmark scores---using models as instruments for understanding behavior. We address this by asking: can we use large vision-language models (LVLMs) as experimental instruments for studying their own failure dynamics? Focusing on visual hallucination, we introduce SAFE, a training-free decoding framework that contrasts visually-grounded and vision-ablated generation paths to produce a token-level contrastive grounding score that identifies when the model favors linguistic priors over visual evidence. This signal serves dual roles: as a practical proxy for detecting visually-ungrounded tokens, and as the basis for decoding-time penalties. Our analysis yields three empirical observations: visual dependency decays over generation, hallucinations co-occur in temporal clusters, and early intervention reduces clustering without substantially degrading fluency. On MMHalBench, SAFE substantially outperforms all compared baselines; results elsewhere are more mixed. We argue that designing contrastive probes exemplifies a broader mission: using models as instruments for scientific understanding. Code: \url{https://github.com/zhaozhipeng1997/SAFE_public}.
\end{abstract}

\section{Introduction}
When strong general-purpose models are increasingly available, a foundational question arises: what should the missions of NLP research be? One answer, which we pursue here, is that progress demands new scientific methodologies---using models as experimental instruments to probe \emph{how} and \emph{why} they behave, rather than treating them only as systems to optimize against static leaderboards.

\begin{figure}[t]
	\centering
	\includegraphics[width=\columnwidth]{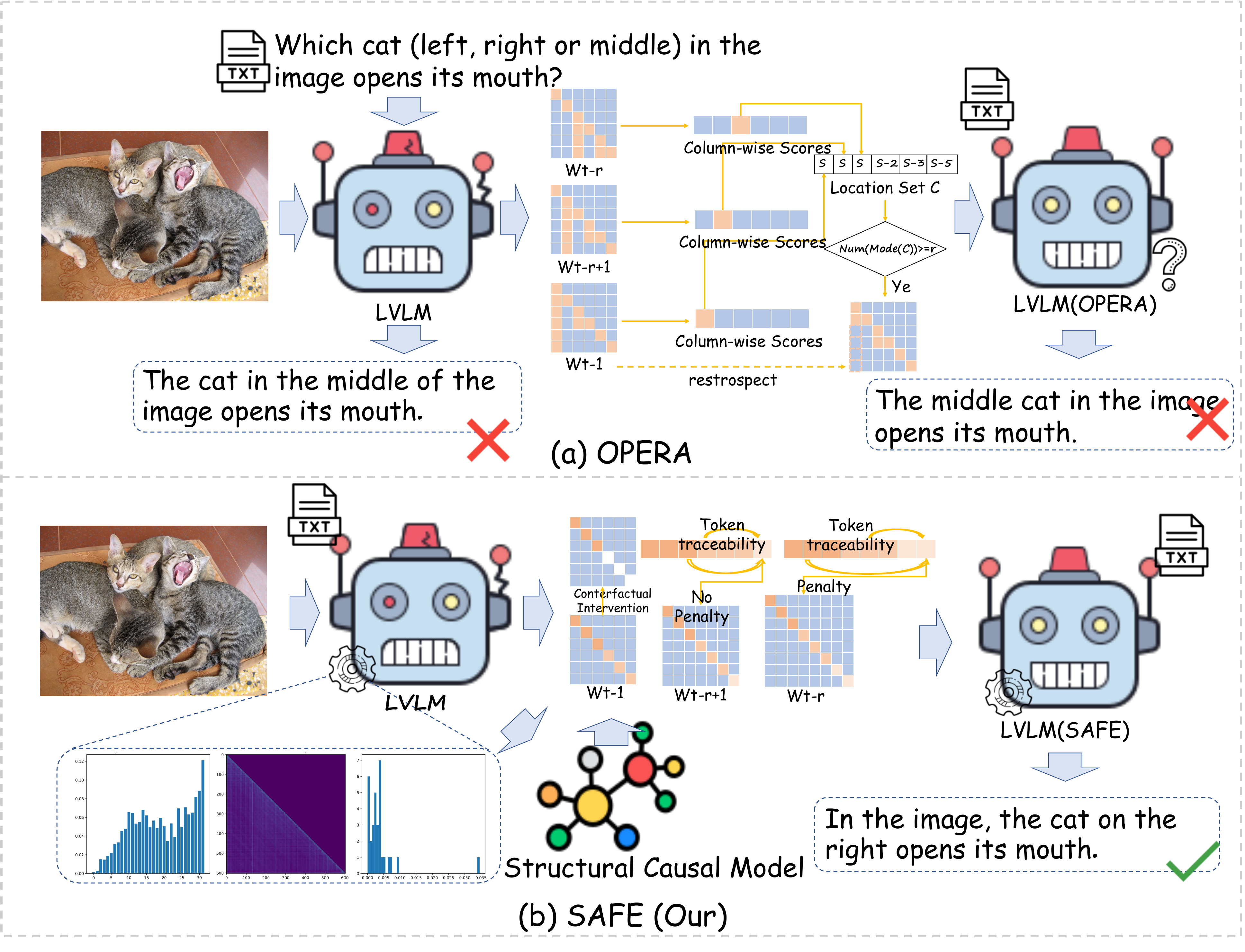} 
	\caption{Comparison of hallucination mitigation paradigms and SAFE.}
	\label{introduction} 
\end{figure}

We apply this perspective to visual-language hallucination in LVLMs \citep{liu2023llava,Qwen2.5VL,wu2024deepseekvl2mixtureofexpertsvisionlanguagemodels}. As illustrated in Figure~\ref{introduction}, even strong models routinely generate text contradicting visual input, eroding trustworthiness.

Current hallucination research is dominated by two paradigms: training-time interventions \citep{chen2025perturbollavareducingmultimodalhallucinations,zhang2024reflectiveinstructiontuningmitigating} and post-hoc decoding heuristics \citep{liu2024payingattentionimagetrainingfree,huang2024opera}. Both improve benchmarks but treat hallucination as a defect to suppress rather than a phenomenon to study. We argue that understanding \emph{how} models negotiate between visual evidence and linguistic priors requires a diagnostic instrument---not another mitigation technique---that quantifies token-level visual grounding and reveals temporal failure dynamics.

We introduce contrastive probing: by running generation paths with and without visual input and computing the token-level log-probability gap, we obtain a \textbf{contrastive grounding score} as a real-time proxy for visual dependency. We emphasize this is a \emph{diagnostic heuristic}, not formal causal identification. We call the framework \textbf{SAFE} (\textbf{S}tructural-\textbf{A}ware \textbf{F}aithfulness \textbf{E}nhancement), operating along two dimensions:
\begin{itemize}
	\item \textbf{As a diagnostic probe:} Token-level visual dependency scores identify which tokens lack grounding and when language priors dominate.
	\item \textbf{As a mitigation method:} SAFE penalizes low-dependency tokens via dual-path contrast with adaptive penalty decay and competitive inhibition.
\end{itemize}
Across five benchmarks and three 7B architectures, visual dependency decays over decoding, hallucinations cluster temporally, and early intervention reduces clustering without substantially degrading fluency. The method is particularly effective on open-ended hallucination benchmarks (MMHalBench), with mixed results where external knowledge or visual ambiguity dominates. This work illustrates a research direction we argue is increasingly central: designing contrastive probes that transform models into instruments for scientific understanding.

\section{Related Work}
LVLMs such as LLaVA-1.5~\citep{liu2023llava}, InstructBLIP~\citep{InstructBLIP}, and Shikra~\citep{chen2023shikra} integrate LLMs~\citep{deepseekai2025deepseekr1incentivizingreasoningcapability,openai2024gpt4technicalreport} with visual encoders, yet all exhibit visual-language hallucinations~\citep{zhang2023sirenssongaiocean,tonmoy2024comprehensive}. This pervasiveness suggests hallucination is rooted in how models balance visual evidence against strong language priors.

\subsection{Hallucination Mitigation: Three Analytical Paradigms}
\label{sec:hal_mitigation}
\paragraph{Training-based alignment.}
Methods fine-tune models on curated data to strengthen visual-semantic correspondence~\citep{chen2025perturbollavareducingmultimodalhallucinations,zhang2024reflectiveinstructiontuningmitigating} or modify architectures for visual feature integration~\citep{xie2024vdpomitigatinghallucinationlarge,shang2024pixelstokensrevisitingobject}. While effective given high-quality data, they obscure \emph{why} hallucination rates decrease and generalize poorly to unseen visual concepts.

\paragraph{Heuristic decoding interventions.}
OPERA~\citep{huang2024opera} penalizes over-reliance on summary tokens. Contrastive decoding~\citep{obrien2023contrastive} has spawned multimodal variants: VCD~\citep{vcd}, ICD~\citep{icd}, ConVis~\citep{park2024convis}, LCD~\citep{manevich2024lcd}, CATCH~\citep{kan2024catch}, IFCD~\citep{zhou2025ifcd}, SDCD~\citep{xia2026sdcd}, ASCD~\citep{wang2025ascd}. SIRA~\citep{qin2026sira} constructs internal counterfactuals via attention masking. DCD~\citep{wang2025mllm} and INTER~\citep{dong2025inter} correct decoding online. Dropout Decoding~\citep{fang2024uncertainty} and ECD~\citep{fieback2025ecd} use uncertainty masking. Calibration by~\citet{fang2025grounding} targets visual-language balance; ClearSight~\citep{yin2025clearsight} amplifies visual signals. Grounding-score methods include VGS-Decoding~\citep{kolli2026vgs} and IECD\textsuperscript{2}~\citep{bangde2026iecd}. CLIP-guided~\citep{deng2024seeing}, GLSim~\citep{park2025glsim}, and HACL~\citep{jiang2024hacl} use auxiliary vision models or contrastive learning. Domain extensions: Med-VCD~\citep{mahdavi2025medvcd}, 3D-VCD~\citep{ogunleye20263dvcd}. SAFE's contribution is the dual-forward-pass log-probability gap with parameterized penalty and diagnostic analysis.

\paragraph{Token-level visual diagnostics.}
VISTA~\citep{li2025hidden} analyzes visual information decay with token-level steering. TruthPrInt~\citep{duan2025truthprint} uses latent truthful signals as per-token indicators.~\citet{cao2026when} and~\citet{nguyen2026beyond} examine attention structure and fine-grained token grounding for hallucination detection. These share SAFE's premise; SAFE's distinction is the log-probability gap between dual passes paired with penalty-based control.

\paragraph{Diagnosis-informed decoding.}
Prior paradigms do not systematically probe \emph{when} visual evidence is abandoned or \emph{how} ungrounded tokens affect generation. Our contrastive probing quantifies token-level visual dependency.~\citet{geigle2024does} show stronger grounding does not always reduce hallucination, contextualizing SAFE's mixed results. The key distinction is that the same signal drives analysis and mitigation within a unified framework.
\begin{figure*}[t]
	\centering
	\includegraphics[width=\textwidth]{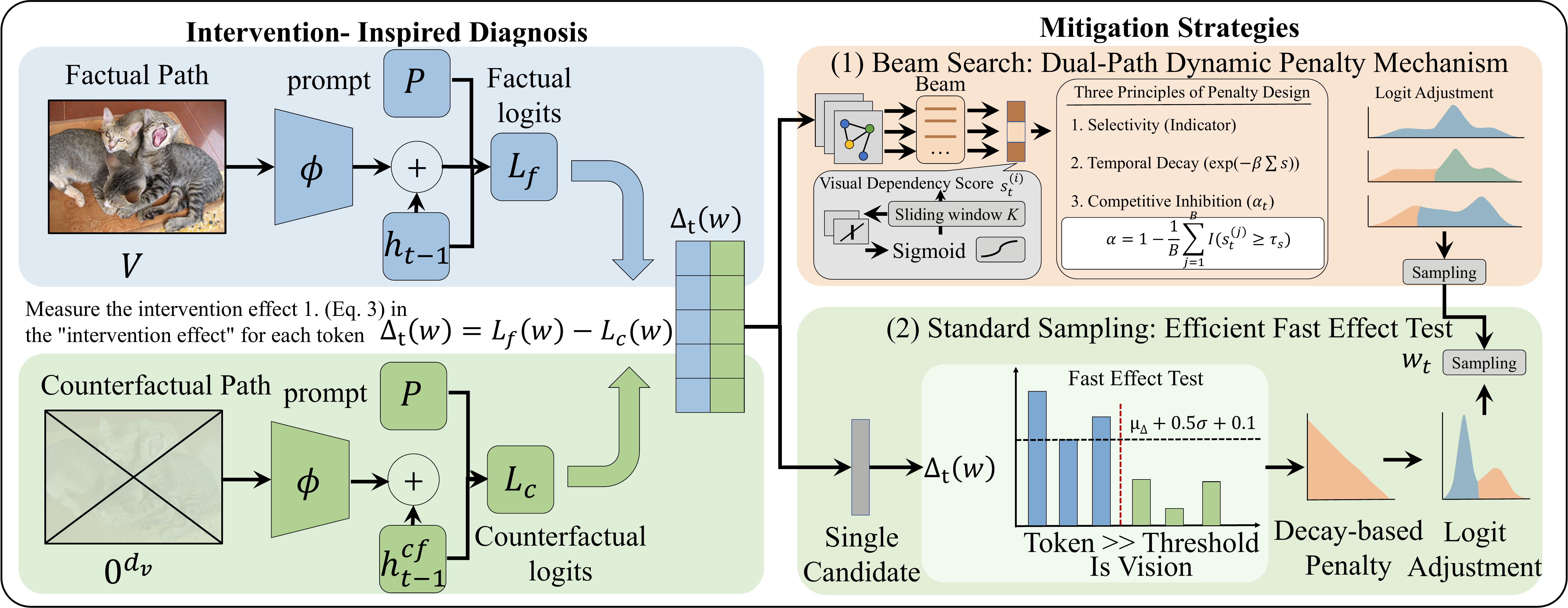} 
	\caption{SAFE dual-path decoding: visually-grounded vs vision-ablated paths with progressive grounding penalty.}
	\label{safe} 
\end{figure*}
\subsection{Empirical Diagnosis via Contrastive Probing}
\label{sec:intervention_inspired}
We connect to controlled manipulation for probing neural models: counterfactual analysis has examined linguistic representations, and perturbation-based methods have studied vision-language relationships---mostly for post-hoc explanation, not real-time diagnosis. We operationalize contrastive probing---comparing behavior with and without a targeted input---as a real-time signal embedded in decoding.
\section{Method}
\subsection{The Central Question: How to Measure a Token's Visual Grounding?}
\label{sec:intervention_concepts}
To what extent does a token depend on visual information versus linguistic priors? Answering this requires isolating visual contributions---a contrastive diagnostic probe. Let the LVLM $\mathcal{M}$ be decomposed into a visual encoder $\phi: \mathcal{V} \to \mathbb{R}^{d_v}$ and language decoder $\psi: \mathcal{P} \to \mathbb{R}^{d_l}$. At time step $t$, the hidden state $h_t \in \mathbb{R}^{d_h}$ evolves as:
\begin{equation}
	h_t = f_\theta \left( [\phi(\mathcal{V}); \psi(\mathcal{P})], h_{t-1} \right)
	\label{eq1}
\end{equation}
where $\phi$ maps image $\mathcal{V}$ to visual features, $\psi$ processes textual prompts $\mathcal{P}$, $[\cdot;\cdot]$ denotes concatenation, and $f_\theta$ is a parameterized Transformer layer.

Equation~\eqref{eq1} cannot separate visual from linguistic contributions. To isolate the visual contribution, we create a diagnostic contrast by zeroing visual features while preserving all other components (see Appendix~\ref{app:causal_diagrams} for conceptual motivation and discussion of alternative ablation strategies). We use $do$-notation descriptively, by analogy to causal formalisms, without implying satisfaction of formal identification conditions. This yields a vision-ablated state:
\begin{equation}
	h_t^{cf} = f_\theta \left( [\mathbf{0}^{d_v}; \psi(\mathcal{P})], h_{t-1}^{cf} \right)
	\label{eq2}
\end{equation}
where $\mathbf{0}^{d_v}$ replaces visual features with a zero vector of the same dimension $d_v$ as $\phi(\mathcal{V})$. The two paths differ \emph{only} in visual input, isolating the visual contribution to token probabilities. We choose zero ablation for determinism, maximal contrast, and preserved linguistic context. Compared to learned null embeddings (MMHalBench 3.48 vs.\ 3.55), shuffled tokens (3.22), and Gaussian noise (3.15), zero ablation performs comparably to learned alternatives while being simpler (full details in Appendix~\ref{app:causal_diagrams}). The \textbf{diagnostic contrast score} is the log-probability difference:
\begin{equation}
	\Delta_t(w) = \log P_{\mathcal{F}}(w|h_t) - \log P_{\mathcal{C}}(w|h_t^{cf})
	\label{eq:intervention_effect}
\end{equation}
where $P_{\mathcal{F}}$ and $P_{\mathcal{C}}$ are visually-grounded and vision-ablated probabilities. $\Delta_t(w) > 0$ means visual information increases the token's probability (visually grounded); $\Delta_t(w) \approx 0$ signals linguistic-prior dominance. This scalar, computed at every step for every candidate, is SAFE's core diagnostic signal. We emphasize that $\Delta_t(w)$ is a diagnostic statistic whose validity rests on its empirical utility for detecting and suppressing hallucinations, not on satisfaction of formal causal identification assumptions.
\subsection{From Diagnosis to Mitigation: The Dual-Path Penalty Mechanism}
\label{sec:dual_path}
$\Delta_t(w)$ provides a per-token grounding estimate. We translate this into mitigation via three principles: sliding-window aggregation for robustness; time-sensitive penalties---strong early, decaying as context accumulates; and global relaxation when most candidates are well-grounded. The penalty is:
\begin{equation}
	s_t^{(i)} = \sigma \left( \frac{1}{\min(K, t)} \sum_{k=\max(1, t-K+1)}^{t} \Delta_k(w^{(i)}) \right)
	\label{eq:visual_dependency}
\end{equation}
where $\sigma(\cdot)$ normalizes to $(0,1)$ and $\min(K,t)$ handles boundaries. We design three penalty properties: (i) \textbf{Selectivity}---only tokens below $\tau_s$ penalized; (ii) \textbf{Temporal decay}---$\exp(-\beta \sum s)$ weakens penalty as visual evidence accumulates; (iii) \textbf{Competitive inhibition}---$\alpha_t$ relaxes when candidates are well-grounded.
\begin{equation}
	\mathcal{P}_t^{(i)} = \lambda \cdot \exp\left(-\beta \sum_{\tau=1}^t s_\tau^{(i)}\right) \cdot \mathbb{I}(s_t^{(i)} < \tau_s)
	\label{eq:penalty}
\end{equation}
where $\lambda$, $\beta$, $\tau_s$ are penalty strength, decay rate, and threshold. The indicator, exponential, and $\alpha_t$ (Equation~\ref{eq:adjusted_logits}) implement the three properties.
The final logit adjustment with competitive inhibition is:
\begin{equation}
	\begin{gathered}
		\log P_{\text{adj}}^{(i)}(w) = \log P_{\mathcal{F}}^{(i)}(w) - \mathcal{P}_t^{(i)} \cdot \alpha_t \\
		\alpha_t = 1 - \frac{1}{B}\sum_{j=1}^B \mathbb{I}(s_t^{(j)} \geq \tau_s)
	\end{gathered}
	\label{eq:adjusted_logits}
\end{equation}
where $\alpha_t$ decreases as more candidates meet the threshold, relaxing the global penalty when visual dependency is already strong. For reproducibility, our default hyperparameters are: window length $K{=}5$, penalty strength $\lambda{=}2.0$, decay rate $\beta{=}0.1$, threshold $\tau_s{=}0.3$, beam size $B{=}3$, and temperature $1.0$. Sensitivity to these choices is reported in Appendix~\ref{sec:ablation}.

\subsection{Two Variants of the Same Diagnostic Principle}
\label{sec:two_variants}
The full mechanism targets beam search. For sampling, a lightweight variant (Section~\ref{Structural-Aware Faithfulness Enhancement}) preserves the dual-path contrast with a fast effect test. Beam uses sliding-window aggregation, exponential decay, and competitive inhibition; sampling uses a single global threshold with simple decay. We treat them as related heuristics within a common framework.

\subsection{Efficient Approximation for Standard Sampling}
\label{Structural-Aware Faithfulness Enhancement}
For standard sampling, we present a lightweight variant preserving the dual-path contrast while replacing the penalty machinery with a fast effect test (see Appendix~\ref{app:causal_diagrams}):
\begin{equation}
	P(Y \mid X \text{ ablated}) = \sum_{Z} P(Y \mid X, Z) P(Z)
\end{equation}
illustrating the contrastive motivation with shared language context $Z$, though exact marginalization is intractable. We approximate by computing $\Delta = \mathbf{L}_f - \mathbf{L}_c$ per step with shared history.
The diagnostic test uses a threshold motivated by Cohen's d:
\begin{equation}
	\text{is\_vision}(w) = \mathbb{I}\left( \Delta(w) > \mu_\Delta + 0.5 \cdot \sigma_\Delta + 0.1 \right)
	\label{eq:fast_effect}
\end{equation}
where $0.5 \cdot \sigma_\Delta$ corresponds to Cohen's d effect size \citep{cohen1988statistical}; $0.1$ prevents false positives when $\sigma_\Delta \approx 0$ (Appendix~\ref{app:threshold_analysis}). Non-visual tokens are penalized analogously:
\begin{equation}
	\mathbf{L}_{\text{adjusted}}(w) = \mathbf{L}(w) - \lambda \cdot \gamma^t \cdot (1 - \text{is\_vision}(w))
\end{equation}
with $\gamma^t = 1/(1+0.1t)$ mirroring temporal decay. This preserves the core principle---measure visual dependency, then intervene---with reduced overhead.
\begin{table*}[t]
	\centering
	\small
	\setlength{\tabcolsep}{5pt}
	\begin{tabular}{lcccccccccccc}
		\toprule
		\multirow{2}{*}{\textbf{Method}} & \multirow{2}{*}{\textbf{qAcc}$\uparrow$} & \multirow{2}{*}{\textbf{fAcc}$\uparrow$} & \multicolumn{3}{c}{\textbf{LV Diagnosis}} & \multicolumn{2}{c}{\textbf{aAcc}} & \textbf{Pct. Diff}&\textbf{FP Ratio} & \multicolumn{3}{c}{\textbf{Consistency}}  \\
		\cmidrule(lr){4-6} \cmidrule(lr){7-8} \cmidrule(lr){9-10} \cmidrule(lr){11-13}
		& & &\textbf{LH$\downarrow$}& \textbf{VI$\downarrow$} &\textbf{Mixed$\downarrow$}& \textbf{Easy}$\uparrow$&\textbf{Hard}$\uparrow$& \textbf{${(\sim0)}$} & \textbf{${(\sim0.5)}$} & \textbf{C}$\uparrow$ & \textbf{I}$\downarrow$ & \textbf{W}$\uparrow$ \\
		\midrule
		Sample &10.76  & 17.91 & 34.96 & 42.40 &22.62  &41.09  &38.60  & 0.29 &0.76&17.91&61.84&20.23 \\
		Beam & 9.01 &15.89  & 36.08 & 41.99 &  21.92& 40.87 & 37.67 & 0.31 &0.77&15.89&63.58&20.52 \\
		OPERA &  11.86& 17.05 & 34.86&  39.14& 25.98 & 42.63 & 41.86 & 0.27 &0.75&17.05&65.31&17.63 \\
		VCD &14.72  & 15.31 & 30.47 & 44.86 & 24.65 & 39.56 &34.65  & 0.24 &0.71&15.31&61.27& 23.41\\
		AGLA & 12.52 &15.02  & 27.81& 49.40 & 22.78 & 36.04& 36.51 & 0.22&0.68&15.02&62.71&22.25 \\
		SID &10.98 & 15.31 & 34.44 &  46.82& 18.73 &38.46 &  34.65& 0.25&0.71&15.31& 60.11&24.56\\
		ICD & 12.96& 16.18 & 31.20 & 44.92 & 23.86 &  41.75& 36.51& 0.24 &  0.71&16.18&62.71&21.09\\
		\textbf{SAFE} &14.50&17.34&35.15&42.50&22.34&41.53&37.90&0.27&0.73&17.34&62.13&20.52 \\
		\bottomrule
	\end{tabular}
	\caption{Performance on HallusionBench using LLaVA-1.5. qAcc: Question Pair Accuracy; fAcc: Figure Accuracy; LH: Language Hallucination; VI: Visual Illusion; C/I/W: Correct/Inconsistent/Wrong.}
	\label{hallusionbench_main}
\end{table*}
\begin{table*}[t]
	\centering
	\small
	\begin{tabular}{lcccccccccccc}
		\toprule
		\multirow{2}{*}{\textbf{Method}} & \multicolumn{4}{c}{\textbf{Random}} & \multicolumn{4}{c}{\textbf{Popular}} & \multicolumn{4}{c}{\textbf{Adversarial}} \\
		\cmidrule(lr){2-5} \cmidrule(lr){6-9} \cmidrule(lr){10-13}
		& \textbf{Acc}$\uparrow$ & \textbf{Prec}$\uparrow$ & \textbf{Recall}$\uparrow$ & \textbf{F1}$\uparrow$ & \textbf{Acc}$\uparrow$ & \textbf{Prec}$\uparrow$ & \textbf{Recall}$\uparrow$ & \textbf{F1}$\uparrow$ & \textbf{Acc}$\uparrow$ & \textbf{Prec}$\uparrow$ & \textbf{Recall}$\uparrow$ & \textbf{F1}$\uparrow$ \\
		\midrule
		Sample & 82.88 & 92.03 & 73.13 & 81.50 & 80.90 & 86.81 & 72.86 & 79.23 & 77.50 & 80.53 & 72.53 & 76.32 \\
		Beam & 86.66&	97.44&	76.13	&85.47	&85.26	&93.14	&76.13&	83.78&	83.63&	89.56&	76.13	&82.30  \\
		OPERA & 88.10 & 95.50 & 80.73 & 87.50 & 85.93 & 90.10 & 80.73 & 85.16 & 82.56 & 83.80 & 80.73 & 82.24 \\
		VCD & 81.75 & 89.81 & 72.86 & 80.45 & 81.73 & 85.68 & 76.20 & 80.66 & 77.26 & 79.21 & 73.93 & 76.48 \\
		AGLA & 82.61 & 92.91 & 71.73 & 80.96 & 82.30 & 88.97 & 73.73 & 80.64 & 78.60 & 82.35 & 72.80 & 77.28 \\
		SID & 83.47 & 85.40 & 81.93 & 83.63 & 80.80 & 79.35 & 83.26 & 81.26 & 74.76 & 71.51 & 82.33 & 76.54 \\
		ICD & 85.87 & 85.65 & 87.20 & 86.42 & 82.46 & 79.87 & 86.80 & 83.19 & 75.83 & 71.27 & 86.53 & 78.16 \\
		\textbf{SAFE} &88.96& 95.94& 82.06& 88.46& 86.46& 89.98& 82.06& 85.84& 82.63& 83.00&82.06 & 82.53	 \\
		\bottomrule
	\end{tabular}
	\caption{POPE results on LLaVA-1.5}
	\label{pope_main}
\end{table*}
\subsection{The Unified SAFE Decoding Algorithm}
\label{SAFE algorithm}

Algorithm~\ref{alg:safe} unifies diagnosis and mitigation: at each step, the diagnostic contrast $\Delta_t$ is computed first, then the penalty mechanism (beam search) or fast effect test (sampling) is applied. The algorithm operates as follows:

\begin{algorithm}[t]
\caption{SAFE: Contrastive Diagnosis and Mitigation for Multimodal Generation}
\label{alg:safe}
\begin{algorithmic}[1]
\REQUIRE Vision-language model $\mathcal{M}$, image $\mathcal{V}$, prompt $\mathcal{P}$, decoding mode $m \in \{\text{beam}, \text{sample}\}$
\ENSURE Generated text sequence $W = w_1, w_2, \dots, w_T$
\STATE Initialize visually-grounded hidden state $h_0$ and vision-ablated hidden state $h_0^{cf}$
\STATE Initialize generated sequence $W \leftarrow \emptyset$
\FOR{$t = 1$ to $T$}
    \STATE Compute visually-grounded logits $\mathbf{L}_f$ using $h_{t-1}$ and visual input $\phi(\mathcal{V})$
    \STATE Compute vision-ablated logits $\mathbf{L}_c$ using $h_{t-1}^{cf}$ and masked visual input $\mathbf{0}^{d_v}$
    \STATE Compute diagnostic contrast $\Delta_t = \mathbf{L}_f - \mathbf{L}_c$

    \IF{$m = \text{beam}$}
        \STATE Apply dual-path dynamic penalty mechanism (Section~\ref{sec:dual_path}):
        \STATE Compute visual dependency scores $s_t^{(i)}$ for each beam candidate
        \STATE Compute penalty terms $\mathcal{P}_t^{(i)}$ and competitive inhibition factor $\alpha_t$
        \STATE Adjust logits: $\text{logit}_{\text{adj}}^{(i)} = \text{logit}_{\mathcal{F}}^{(i)} - \mathcal{P}_t^{(i)} \cdot \alpha_t$
    \ELSE
        \STATE \text{(standard sampling)} Apply structural-aware faithfulness enhancement (Section~\ref{Structural-Aware Faithfulness Enhancement}):
        \STATE Compute fast effect test for each token $w$: $\text{is\_vision}(w) = \mathbb{I}(\Delta_t(w) > \mu_\Delta + 0.5 \cdot \sigma_\Delta + 0.1)$
        \STATE Adjust logits for each token $w$: $\mathbf{L}_{\text{adjusted}}(w) = \mathbf{L}_f(w) - \lambda \cdot \gamma^t \cdot (1 - \text{is\_vision}(w))$
    \ENDIF

    \STATE Sample next token $w_t$ from adjusted logits distribution
    \STATE Append $w_t$ to $W$
    \STATE Update hidden states $h_t$ and $h_t^{cf}$ for next step
\ENDFOR
\RETURN $W$
\end{algorithmic}
\end{algorithm}

\section{Experiment}
We ask two questions: do SAFE's diagnostic signals improve benchmark performance, and what patterns do internal signals reveal? We validate on benchmarks (Section~\ref{sec:benchmark_validation}), then analyze token-level dependency, temporal clustering, and fluency (Sections~\ref{sec:token_dep}--\ref{sec:text_quality}). Default hyperparameters ($K{=}5$, $\lambda{=}2.0$, $\beta{=}0.1$, $\tau_s{=}0.3$, $B{=}3$, temperature $1.0$) are selected on a validation split and used uniformly across all models and benchmarks; sensitivity analysis is in the appendix. SAFE incurs $\sim$$2\times$ the inference cost of comparable methods (Table~\ref{time_analysis}, Limitations).
\begin{table*}[t]
	\centering
	\small
	\begin{tabular}{lccccccccccc}
		\toprule
		\multirow{2}{*}{\textbf{Method}} & \multirow{2}{*}{\textbf{Overall}$\uparrow$} & \multirow{2}{*}{\textbf{HR}$\downarrow$} & \multicolumn{8}{c}{\textbf{Categories}} \\
		\cmidrule(lr){4-11}
		& & & \textbf{Attri}$\uparrow$ & \textbf{Adver}$\uparrow$ & \textbf{Comp}$\uparrow$ & \textbf{Counting} $\uparrow$& \textbf{Relation}$\uparrow$ & \textbf{Env}$\uparrow$ & \textbf{Holistic}$\uparrow$ & \textbf{Other}$\uparrow$ \\
		\midrule
		Sample & 1.54 & 0.75 & 2.00 & 0.17 & 2.33 & 1.42 & 1.92 & 2.08 & 0.58 & 1.83 \\
		Beam & 1.49	&0.76&	1.83&	1.17&	1.42&	1.83&	2.17&	1.58&	1.08&	0.83 \\
		OPERA & 1.69 & 0.74 & 2.17 & 1.17 & 2.25 & 1.83 & 1.25 & 2.08 & 1.42 & 1.33 \\
		VCD & 1.33 & 0.83 & 1.08 & 1.50 & 2.00 & 1.00 & 1.58 & 1.33 & 0.83 & 1.33 \\
		AGLA & 1.32 & 0.82 & 1.83 & 1.42 & 0.92 & 1.00 & 1.50 & 2.08 & 0.75 & 1.08 \\
		SID & 1.34 & 0.80 & 1.75 & 1.58 & 2.25 & 0.83 & 1.08 & 1.08 & 0.58 & 1.58 \\
		ICD & 1.41 & 0.81 & 2.17 & 1.00 & 1.17 & 1.67 & 1.75 & 1.75 & 0.58 & 1.17 \\
		\textbf{SAFE} &3.55 & 0.44 & 3.25 & 3.17 & 3.83 &3.08 & 4.25 & 4.42 & 3.42 &3.0  \\
		\bottomrule
	\end{tabular}
	\caption{Performance on MMHalBench using LLaVA-1.5. HR: Hallucination Ratio; Attri: Attribute; Adver: Adversarial; Comp: Comparison; Env: Environment.}
	\label{mmhalbench_main}
\end{table*}
\begin{figure*}[t]
	\centering
	\includegraphics[width=\textwidth]{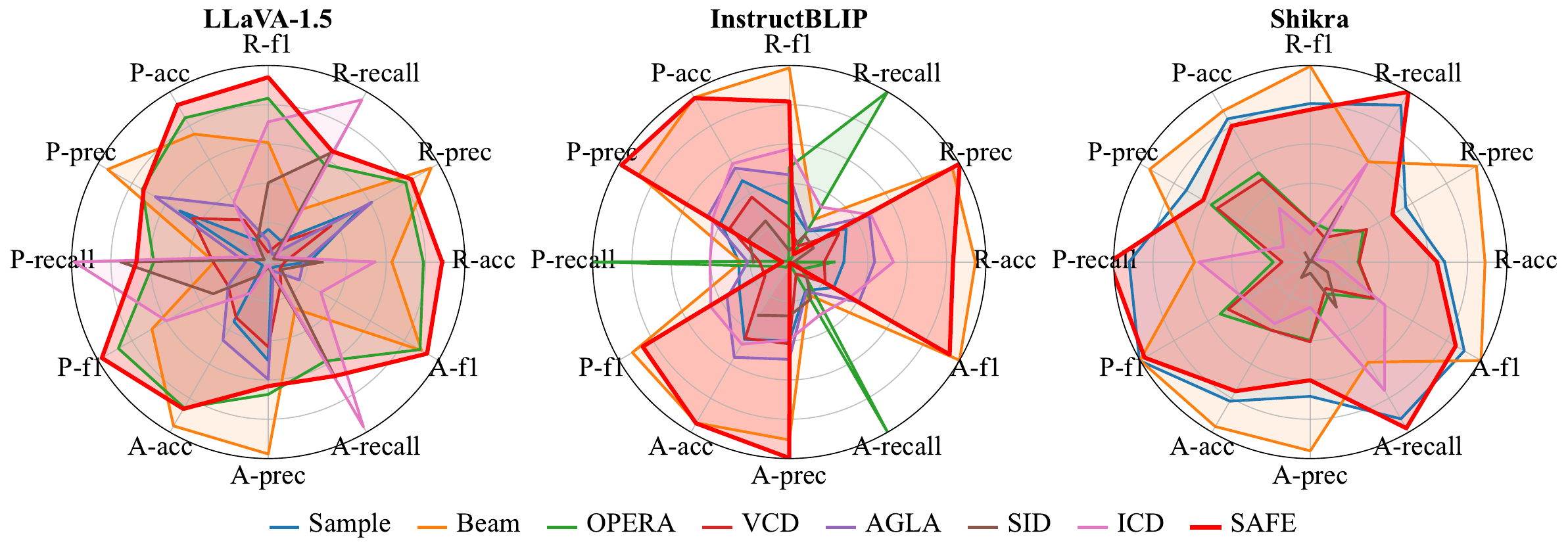} 
	\caption{POPE results across models. R/P/A: Random/Popular/Adversarial.}
	\label{pope_3models} 
\end{figure*}
\subsection{Benchmarks and Experimental Setup}
We evaluate on five benchmarks: HallusionBench~\citep{Guan_2024_CVPR} (language hallucinations and visual illusions), MMHalBench~\citep{sun2023aligninglargemultimodalmodels} (open-ended hallucination), CHAIR~\citep{rohrbach-etal-2018-object} (object hallucination at sentence/instance levels), POPE~\citep{li2023evaluatingobjecthallucinationlarge} (object existence), and MMMU~\citep{yue2023mmmu} (multimodal reasoning). Beam search for MMHalBench, CHAIR, POPE; sampling for HallusionBench and MMMU. Baselines: Sample, Beam, OPERA~\citep{huang2024opera}, VCD~\citep{vcd}, AGLA~\citep{an2024agla}, SID~\citep{huo2025selfintrospectivedecodingalleviatinghallucinations}, ICD~\citep{icd} on LLaVA-1.5~\citep{liu2023llava}, InstructBLIP~\citep{InstructBLIP}, Shikra~\citep{chen2023shikra} (all 7B).
\begin{figure*}[t]
	\centering
	\includegraphics[width=\textwidth]{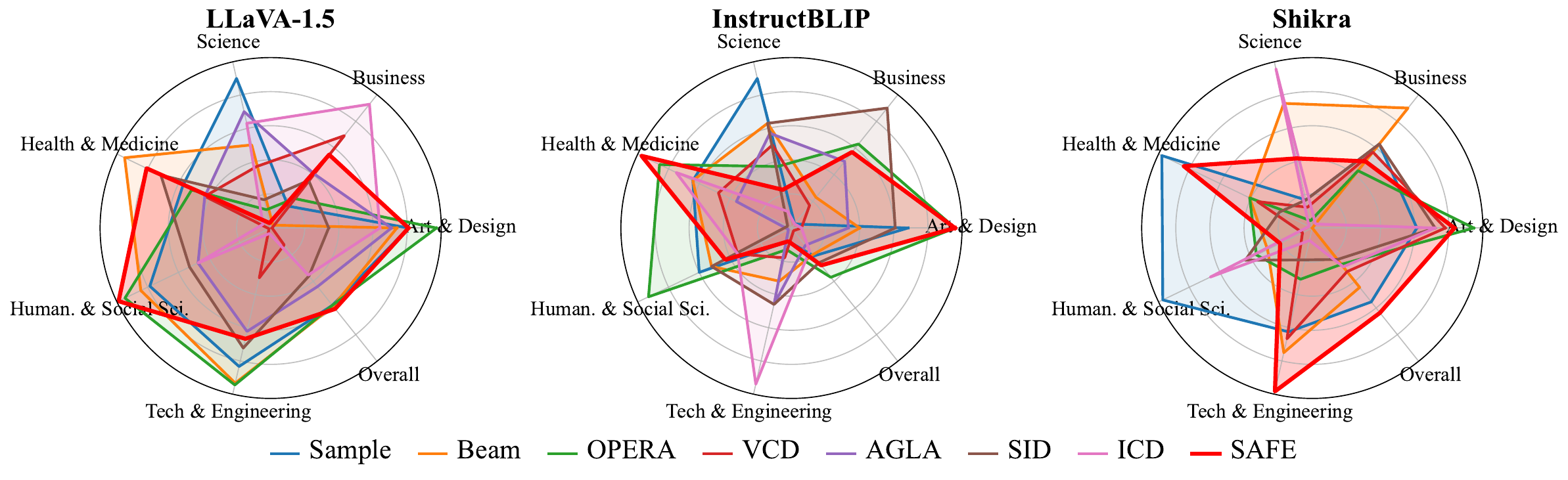} 
	\caption{MMMU results across models.}
	\label{mmmu_3models} 
\end{figure*}
\subsection{Results}
\label{sec:benchmark_validation}
We test whether SAFE's detected patterns correspond to standard benchmark penalties. All experiments: LLaVA-1.5, InstructBLIP, Shikra (7B), temperature 1.0, beam size 3.
\textbf{HallusionBench.} HallusionBench includes visual illusions where visual reliance can produce errors, making ``visual dependency'' ambiguous. SAFE achieves competitive qAcc (14.50) and fAcc (17.34), but VCD has higher qAcc (14.72) and OPERA comparable fAcc (17.05). This is expected: dependency diagnosis works best with reliable visual evidence.

\noindent\textbf{MMHalBench.} SAFE achieves overall score 3.55---more than double the nearest baseline (OPERA, 1.69)---with hallucination ratio 0.44 versus 0.74--0.83 (Table~\ref{mmhalbench_main}). Gains are largest in Relation (4.25) and Environment (4.42), categories penalizing visually inconsistent claims about spatial and contextual relationships---precisely SAFE's target. Unlike CHAIR (object-level string matching) or POPE (binary questions), MMHalBench uses GPT-based evaluation of free-form answers, making it sensitive to visual-semantic consistency across multi-token descriptions. This gain warrants caution: MMHalBench relies on a single GPT-based judge over 96 samples, and SAFE's conservative generation may be scored favorably in this rubric. We treat this as strong but benchmark-specific evidence.
\begin{table*}[t]
	\centering
	\small
	\begin{tabular}{lccccccc}
		\toprule
		\multirow{2}{*}{\textbf{Method}} & \textbf{Art \&} & \multirow{2}{*}{\textbf{Business}$\uparrow$}  & \multirow{2}{*}{\textbf{Science}$\uparrow$} & \textbf{Health \&} & \textbf{Human. \&} & \textbf{Tech \& } & \multirow{2}{*}{\textbf{Overall}$\uparrow$} \\
		& \textbf{Design}$\uparrow$ &  &  & \textbf{Medicine}$\uparrow$ & \textbf{Social Sci.}$\uparrow$ & \textbf{Eng.}$\uparrow$ & \\
		\midrule
		Sample & 0.492 & 0.213 & 0.320 & 0.300 & 0.458 & 0.305 & 0.337 \\
		Beam & 0.475&	0.193&	0.28&	0.353&	0.475	&0.314&	0.338 \\
		OPERA & 0.533 & 0.221 & 0.241 & 0.292 & 0.505 & 0.315 & 0.337 \\
		VCD & 0.292 & 0.287 & 0.267 & 0.280 & 0.233 & 0.257 & 0.269 \\
		AGLA & 0.467 & 0.247 & 0.300 & 0.280 & 0.367 & 0.286 & 0.316 \\
		SID & 0.375 & 0.240 & 0.247 & 0.320 & 0.383 & 0.295 & 0.304 \\
		ICD & 0.450 & 0.320 & 0.293 & 0.227 & 0.367 & 0.233 & 0.303 \\
		\textbf{SAFE} & 0.492&0.267&0.233&0.333&0.517&0.29&0.341	\\
		\bottomrule
	\end{tabular}
	\caption{MMMU results on LLaVA-1.5}
	\label{mmmu_main}
\end{table*}

\noindent\textbf{CHAIR.} SAFE achieves CHAIRi 14.8 (competitive with OPERA's 14.5) and recall 74.6 (Table~\ref{chair}), but OPERA is better on CHAIRs (50.5 vs.\ 53.0). On Shikra, SAFE ties best CHAIRi (13.8) but worst CHAIRs (60.0) (Appendix Tables~\ref{appendix_instructblip_chair}--\ref{appendix_shikra_chair}). Instance-level strength is expected---token diagnosis detects unsupported objects---while sentence-level gap reflects discourse planning beyond per-token scope. SAFE generates shorter captions (e.g., 95.3 vs.\ 101.8 for LLaVA-1.5); examined in Section~\ref{sec:token_dep}.
\begin{table}[t]
	\centering
	\small
	\begin{tabular}{lcccc}
		\toprule
		\textbf{Method} & \textbf{CHAIRs$\downarrow$} & \textbf{CHAIRi$\downarrow$} & \textbf{Recall$\uparrow$} & \textbf{Len} \\
		\midrule
		Sample & 58.3 & 17.8 & 69.8 & 101.8 \\
		Beam & 56.2 & 16.2 & 76.4 & 102.6 \\
		OPERA & \textbf{50.5} & \textbf{14.5} & 76.1 & 91.9 \\
		VCD & 51.6 & 15.1 & 70.9 & 103.5 \\
		AGLA & 56.2 & 16.4 & 70.7 & 96.5 \\
		SID & 52.5 & 16.3 & 65.6 & 91.9 \\
		ICD & 56.5 & 16.4 & 72.8 & 94.0 \\
		SAFE & 53.0& 14.8 & 74.6 &95.3  \\
		\bottomrule
	\end{tabular}
	\caption{Performance on CHAIR using LLaVA-1.5. CHAIRs/CHAIRi: sentence/instance-level hallucination ($\downarrow$); Recall: correct objects ($\uparrow$); Len: average caption length.}
	\label{chair}
\end{table}
\begin{table}
	\centering
	\small
	\begin{tabular}{cccccc}
		\toprule
		\textbf{}&\textbf{Setting}&\textbf{TNS}& \textbf{TNT$_\downarrow$} & \textbf{NVAT$_\uparrow$} &\textbf{ARVAT$_\uparrow$}\\
		\midrule
		\multirow{2}{*}{HB}&w/o SAFE &951 &31814&	1727&	5.43\%		\\
		&w/ SAFE &951 &31778&	\textbf{1757}&	\textbf{5.53\%}		\\
		\multirow{2}{*}{MB}&w/o SAFE &96 &3715&42&1.13\%		\\
		&w/ SAFE &96 &3546&\textbf{50}	&\textbf{1.41\%	}	\\
		\bottomrule
	\end{tabular}
	\caption{Token Dependency Analysis in Multimodal Models. Abbreviations: TNS (Total Samples), TNT (Total Tokens), NVAT (Visually Associated Tokens), ARVAT (Avg. Ratio of Visually Associated Tokens). Identifiers: HB (HallusionBench), MB (MMHalBench). Setting indicates with (w/) or without (w/o) SAFE.}
	\label{DependencyAnalysis}
\end{table}

\noindent\textbf{POPE.} SAFE achieves highest precision (95.94 Random, 89.98 Popular) with stable Recall (82.06), confirming the penalty targets visually unsupported tokens. Adversarial results (83.00/82.06) mirror this balanced pattern. See Figure~\ref{pope_3models}.
\begin{table}[t]
	\centering
	\small
	\begin{tabular}{ccccc}
		\toprule
		\textbf{Benchmarks}&\textbf{Decode}&\textbf{PPL1$_{\downarrow}$}& \textbf{PPL2$_{\downarrow}$}\\
		\midrule
		\multirow{4}{*}{HallusionBench} &Sample & 13.0661 & 34.9492 \\
		& Beam & 12.7794 & 34.1806 \\
		& OPERA & 16.9347 &  31.6977 \\
		& SAFE & 12.9545 & 34.0812 \\
		\midrule
		\multirow{4}{*}{MMHalBench} &Sample & 11.7508 &27.7578  \\
		& Beam &10.4976 &25.3651  \\
		& OPERA & 45.2302 & 59.1796  \\
		& SAFE & 10.2301 & 24.7059 \\
		\bottomrule
	\end{tabular}
	\caption{Quality Assessment of Generated Texts}
	\label{textqulity}
\end{table}

\noindent\textbf{MMMU.} MMMU reveals the \emph{boundary} of visual dependency diagnosis: SAFE's overall score (0.341 vs.\ Beam 0.338) is marginally higher but inconsistent. Humanities \& Social Sciences (0.517) benefits from visual interpretation, while Science (0.233 vs.\ Sample 0.320) shows the method cannot help when answers require non-visual knowledge. MMMU illustrates a clear domain boundary. See Figure~\ref{mmmu_3models} and appendix.
\begin{table}
	\centering
	\small
	\begin{tabular}{cccc}
		\toprule
		\textbf{Benchmarks}&\textbf{Method}&\textbf{$P_{\text{prop}}$}&\textbf{$N_{\text{primary}}$}\\
		\midrule
		\multirow{2}{*}{HB}&Single-window & 0.2929 & 454 \\
		&SAFE & 0.1816$_{\downarrow38.0\%}$ &457 \\
		\multirow{2}{*}{MB}&Single-window & 0.0555 & 18 \\
		&SAFE & 0.0454$_{\downarrow18.2\%}$ &22 \\
		\bottomrule
	\end{tabular}
	\caption{Evaluation of Propagative Hallucination. $P_{\text{prop}}$ denotes propagative hallucination probability (the likelihood that an error at step $t$ is followed by errors in subsequent steps), while $N_{\text{primary}}$ represents the count of antecedent errors.}
	\label{chuanbo}
\end{table}
\begin{figure*}[t]
    \centering
    \includegraphics[width=\textwidth]{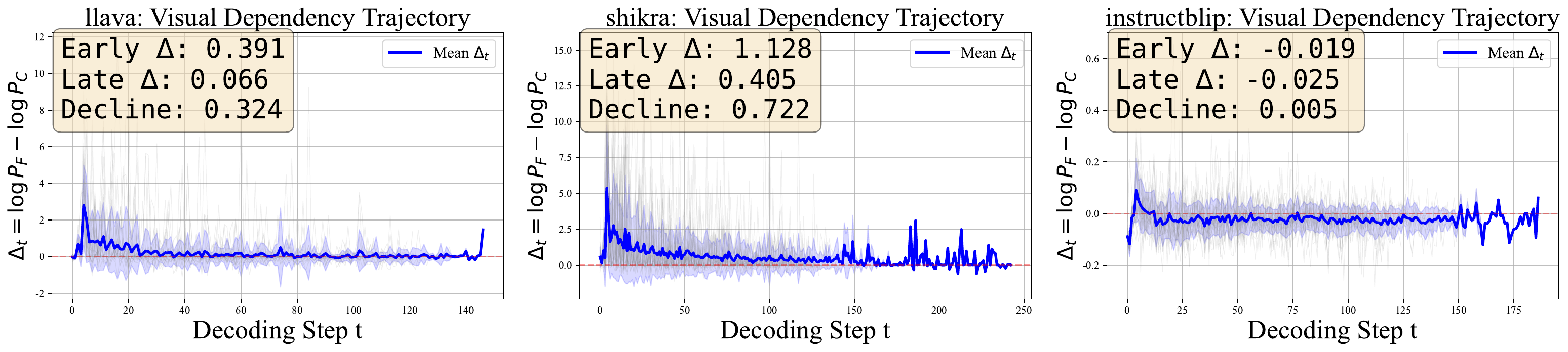}
    \caption{$\Delta_t$ trajectories on 30 COCO images. Gray: individual samples; blue: mean $\pm$1 std; red dashed: $\Delta_t=0$. LLaVA $\Delta_t$ declines from 0.39 (early) to 0.07 (late); Shikra from 1.13 to 0.41.}
    \label{fig:diagnostic_trajectory}
\end{figure*}
\subsubsection{Token-Level Visual Dependency Analysis}
\label{sec:token_dep}
Table~\ref{DependencyAnalysis} reports visually associated tokens---tokens where $\Delta_t$ indicates visual grounding---with and without SAFE. SAFE increases the proportion of visually grounded tokens (HB: 5.43\%$\to$5.53\%; MB: 1.13\%$\to$1.41\%) while total tokens \emph{decreases} (HB: 31,814$\to$31,778; MB: 3,715$\to$3,546). This suggests SAFE selectively suppresses low-dependency tokens, yielding shorter, more visually-attentive outputs.

To verify the ``visual dependency decay'' claim, we collected per-step $\Delta_t$ trajectories via dual-path inference on COCO images. Figure~\ref{fig:diagnostic_trajectory} shows aggregate trajectories: LLaVA $\Delta_t$ declines from 0.39 (first half) to 0.07 (second half); Shikra from 1.13 to 0.41, providing direct evidence that visual grounding weakens as language priors accumulate.\footnote{InstructBLIP's Q-Former produces $\Delta_t{\approx}0$ because learned queries dominate cross-modal fusion; trajectories are informative for LLaVA and Shikra.}

\paragraph{Length-controlled analysis.}
To address the concern that SAFE's gains may reflect output shortening, we compute length-normalized CHAIRs$_{\text{norm}} = \text{CHAIRs} / \text{mean caption length}$ and hallucination rate per 100 tokens. On LLaVA-1.5, SAFE achieves CHAIRs$_{\text{norm}}{=}0.60\%$ (Beam 0.62\%, Sample 0.65\%), 16.0 hallucinated tokens/100 (Beam 16.2, Sample 17.8). Improvement direction persists after normalization, though margins are modest. On Shikra, OPERA (0.59\%) and VCD (0.60\%) outperform SAFE (0.74\%) on CHAIRs$_{\text{norm}}$, consistent with weaker sentence-level performance on this architecture. Full results in Appendix~\ref{app:length_control}.

\subsubsection{Validating the Diagnostic Signal}
\label{sec:signal_validation}
\begin{figure}[t]
    \centering
    \includegraphics[width=\columnwidth]{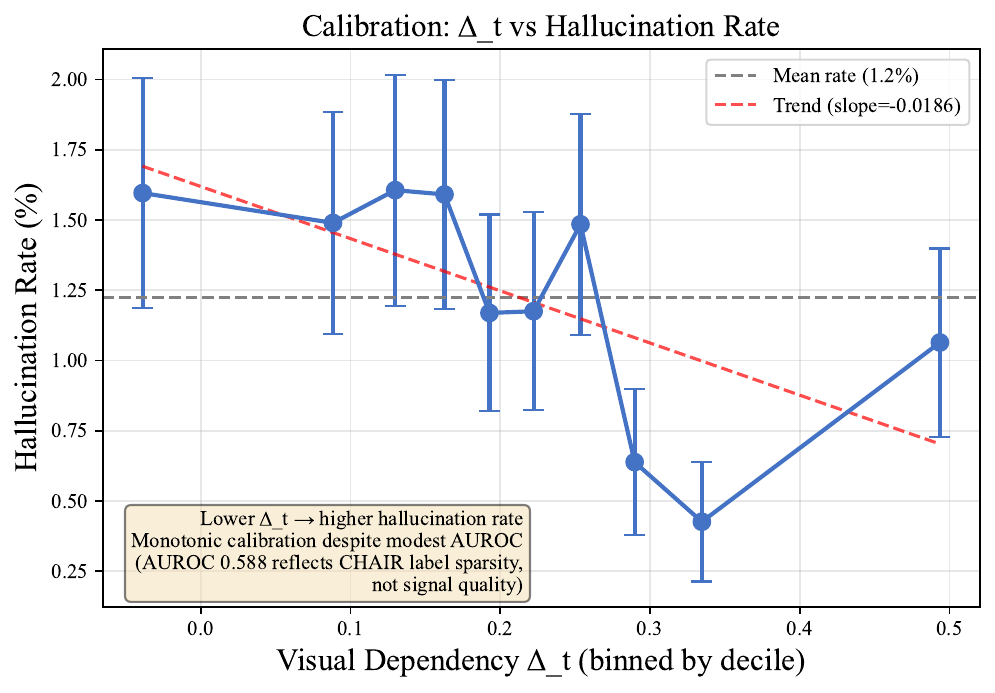}
    \caption{Calibration of $\Delta_t$ against CHAIR hallucination labels. Tokens binned by $\Delta_t$ decile; shaded band shows $\pm1$ standard error. Hallucination rate declines monotonically as $\Delta_t$ increases (trend slope $-0.019$), confirming the signal is well-calibrated despite modest absolute AUROC.}
    \label{fig:calibration}
\end{figure}

To verify that $\Delta_t$ captures grounding-specific information, we compare it against token-level entropy $H$ and confidence $\max p(w)$ for predicting CHAIR hallucination labels on 50 COCO images (same beam-search sequences). $\Delta_t$ achieves AUROC $0.588$, outperforming entropy ($0.380$) and confidence ($0.401$), both of which fall \emph{below} random ($0.50$). A calibration analysis confirms reliability: hallucination rate decreases monotonically from $1.60\%$ (lowest $\Delta_t$ decile) to $0.43\%$ (ninth decile, slope $-0.019$, Figure~\ref{fig:calibration}). The modest absolute AUROC reflects CHAIR's annotation sparsity (${\sim}1.2\%$ object-level labels) rather than poor signal quality. On 100 COCO images (9,397 pairs), grounded $\Delta_t{=}0.211$ vs.\ hallucinated $0.185$ (AUROC $0.574$). Sentence-level aggregation fails (AUROC $0.464$, Appendix~\ref{app:signal_validation}).

\subsubsection{Does Suppressing Hallucinations Compromise Fluency?}
\label{sec:text_quality}
Table~\ref{textqulity} reports perplexity under Qwen2-7B and GPT-2-medium. SAFE achieves lower perplexity than Sample and Beam on MMHalBench (PPL1: 10.23 vs.\ 11.75/10.50), competitive on HallusionBench. This is tentative evidence that removing ungrounded tokens improves coherence, though external LM perplexity is a rough proxy for multimodal outputs. Stronger claims require human evaluation.
\subsubsection{The Chain Structure of Hallucinations}
\label{sec:propagative}
Table~\ref{chuanbo} compares SAFE against a single-window baseline on propagative hallucination probability. SAFE reduces temporal clustering by 38.0\% (HallusionBench) and 18.2\% (MMHalBench). We caution $P_{\text{prop}}$ measures co-occurrence, not causal propagation. SAFE's sliding window detects low-dependency clusters, and the cumulative penalty provides escalating pressure. See Appendix~\ref{CaseStudy}.
\section{Conclusion}
This paper asked: \emph{what are the missions of NLP research when strong models are widely available?} Our answer, via SAFE, is that designing instruments to probe \emph{how} models work---not just measuring \emph{whether} they work---is a central mission. Token-level contrastive probing reveals that visual grounding decays over generation, hallucinations co-occur temporally, and early intervention reduces clustering without substantially degrading fluency. The method is effective on open-ended hallucination (MMHalBench), with mixed results elsewhere---informing \emph{when} token-level diagnosis helps. SAFE demonstrates that using models as experimental instruments yields insights complementing architectural innovation and benchmark optimization, representing a productive direction for NLP.
\section*{Limitations}
SAFE incurs $\sim$$2\times$ the inference cost of comparable methods (Table~\ref{time_analysis}); this could be reduced through shared KV-caching between paths. $\Delta_t(w)$ captures association between visual input and token probability rather than a formal causal mechanism; its interpretation as ``visual dependency'' is validated by empirical hallucination reduction, not causal proofs. The visual ablation (zeroing) assumes clean visual-linguistic separation, weakened under signal entanglement (e.g., text in images). Our analysis derives from general-domain benchmarks and may differ in specialized domains. Benchmark gains do not justify deployment in high-stakes settings such as medical diagnosis. SAFE's conservative decoding may suppress useful details---harmful in assistive applications where informativeness matters---and this faithfulness-informativeness tradeoff may disproportionately underspecify culturally specific or low-frequency content. We use CHAIR and POPE recall and external-LM perplexity as indirect informativeness proxies; a dedicated coverage and human-informativeness study is left for future work. Finally, MMHalBench scores rely on a single GPT-based judge, and we have not fully disentangled hallucination reduction from the confound of shorter output length; multi-judge or human validation remains future work.


\bibliography{custom}

\appendix

\begin{figure*}
	\centering
	\includegraphics[width=\textwidth]{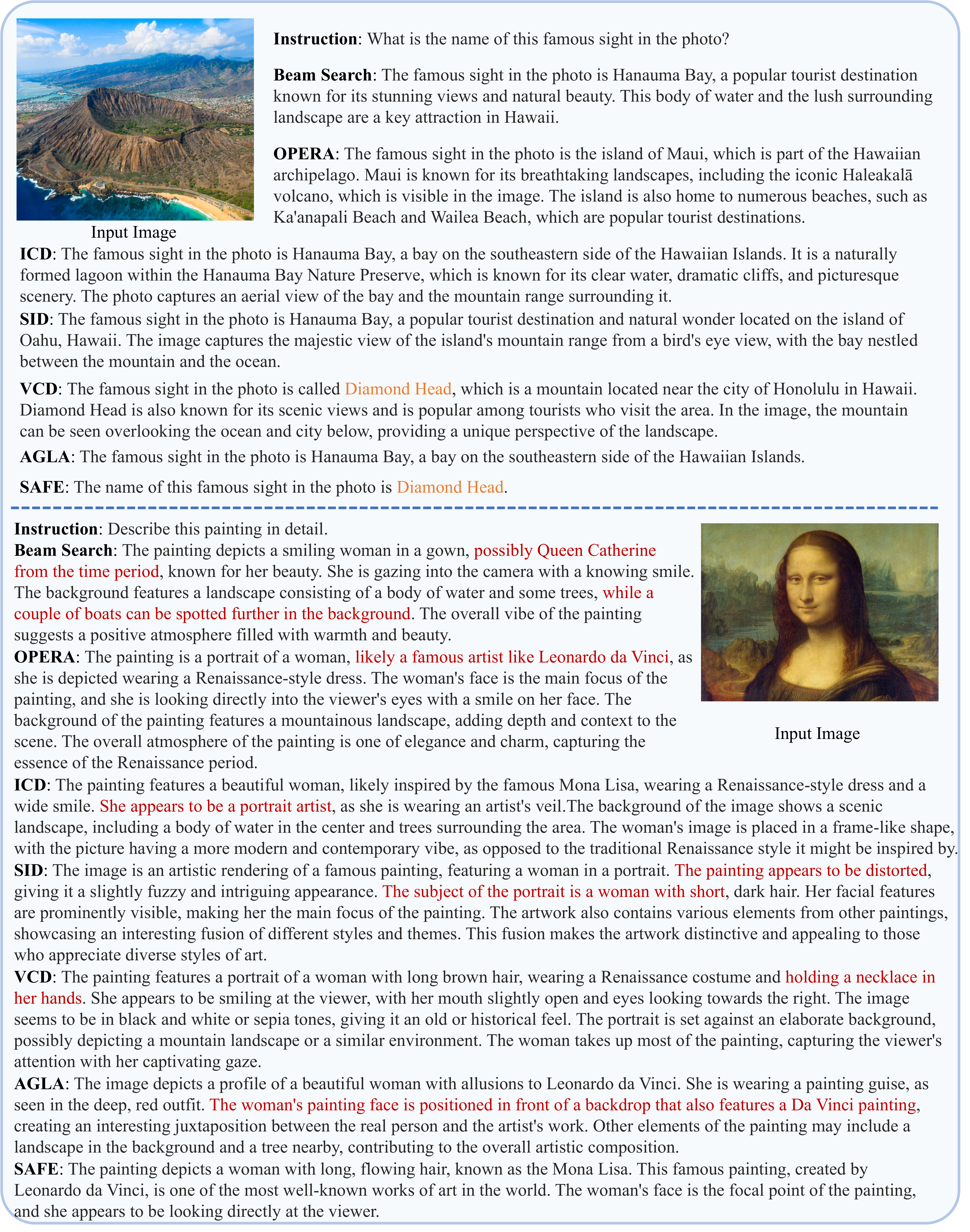} 
	\caption{SAFE’s performance on reducing hallucinations of InstructBLIP. Hallucinations are marked in red. The key word is orange.}
	\label{instructblip_case} 
\end{figure*}
\begin{figure*}
	\centering
	\includegraphics[width=\textwidth]{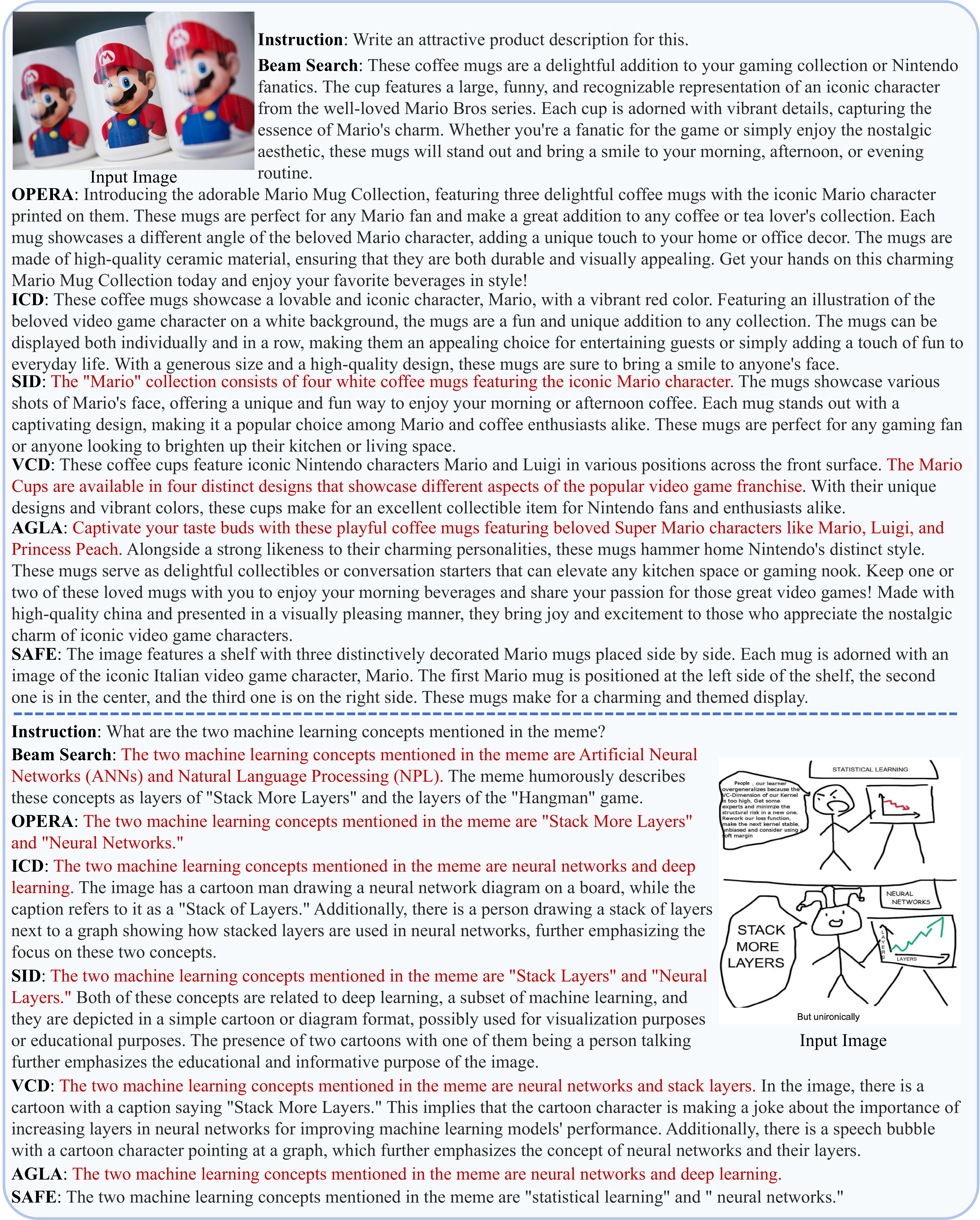} 
	\caption{SAFE’s performance on reducing hallucinations of Shikra. Hallucinations are marked in red.}
	\label{shikra_case} 
\end{figure*}
\begin{figure*}
	\centering
	\includegraphics[width=\textwidth]{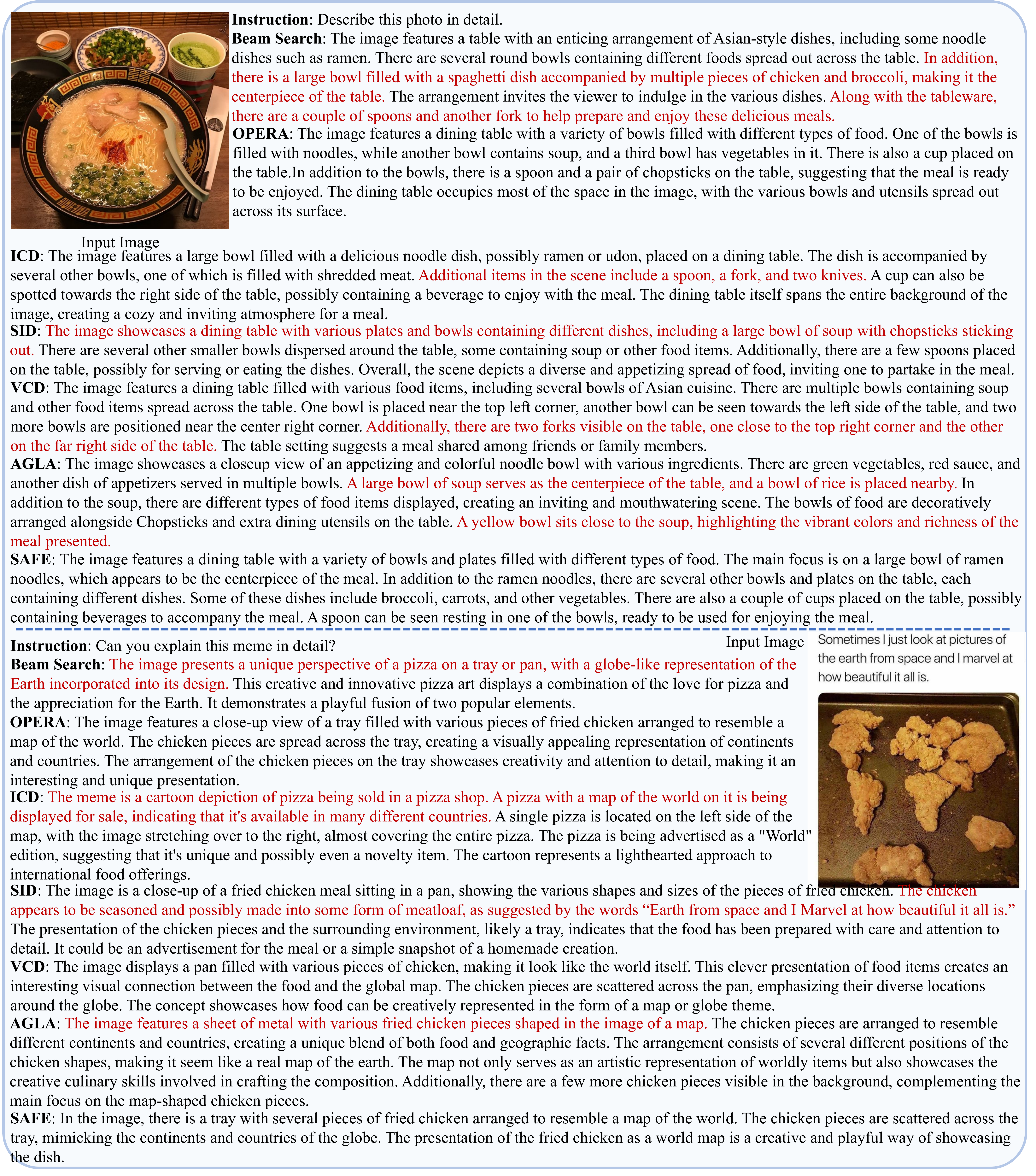} 
	\caption{SAFE’s performance on reducing hallucinations of LLaVA-1.5. Hallucinations are marked in red.}
	\label{llava_case} 
\end{figure*}
\section{Length-Controlled CHAIR Results}\label{app:length_control}
\begin{figure*}[t]
    \centering
    \includegraphics[width=\textwidth]{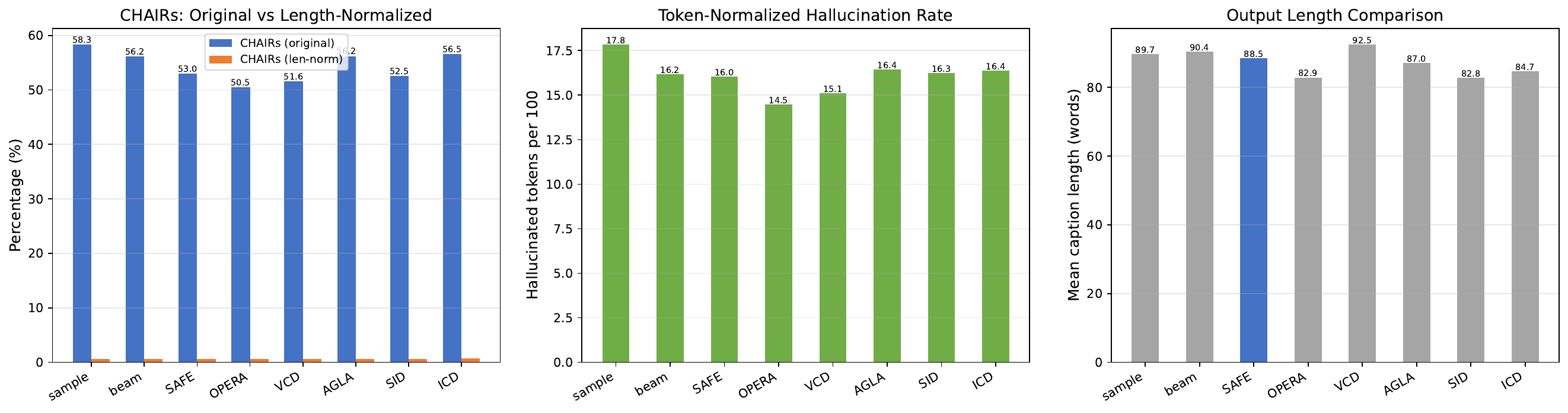}
    \caption{Length-controlled CHAIR analysis for LLaVA-1.5. Left: CHAIRs vs.\ length-normalized CHAIRs. Center: hallucination rate per 100 tokens. Right: mean caption length. SAFE's advantage persists after normalization (CHAIRs$_{\text{norm}}$ 0.60\% vs.\ Beam 0.62\%).}
    \label{fig:length_control_llava}
\end{figure*}

\begin{figure*}[t]
    \centering
    \includegraphics[width=\textwidth]{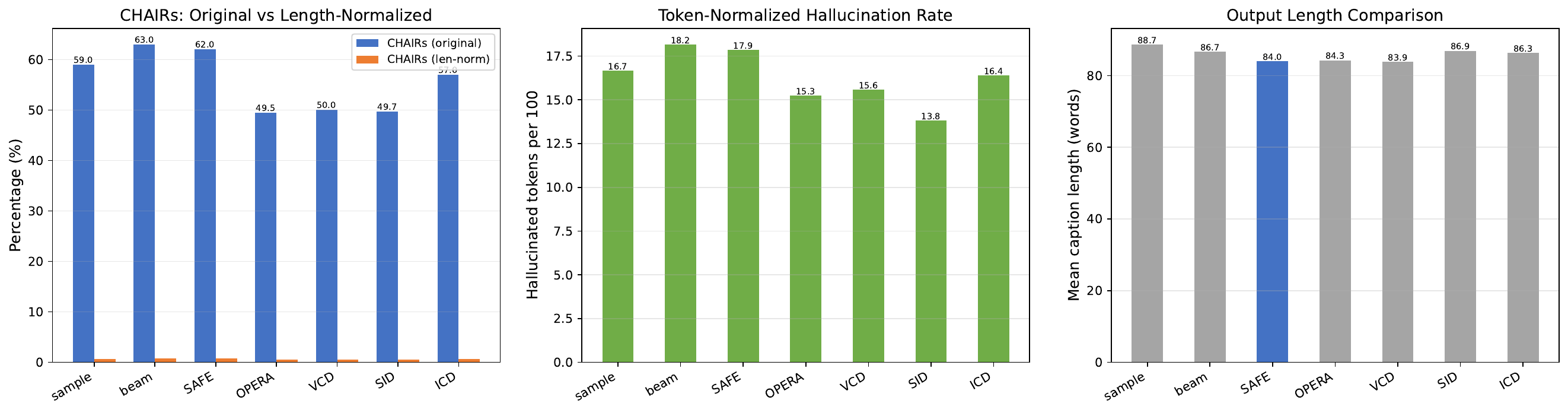}
    \caption{Length-controlled CHAIR analysis for Shikra. OPERA (0.59\%) and VCD (0.60\%) outperform SAFE (0.74\%) on CHAIRs$_{\text{norm}}$, consistent with SAFE's weaker sentence-level performance on this architecture.}
    \label{fig:length_control_shikra}
\end{figure*}

\begin{figure*}[t]
    \centering
    \includegraphics[width=\textwidth]{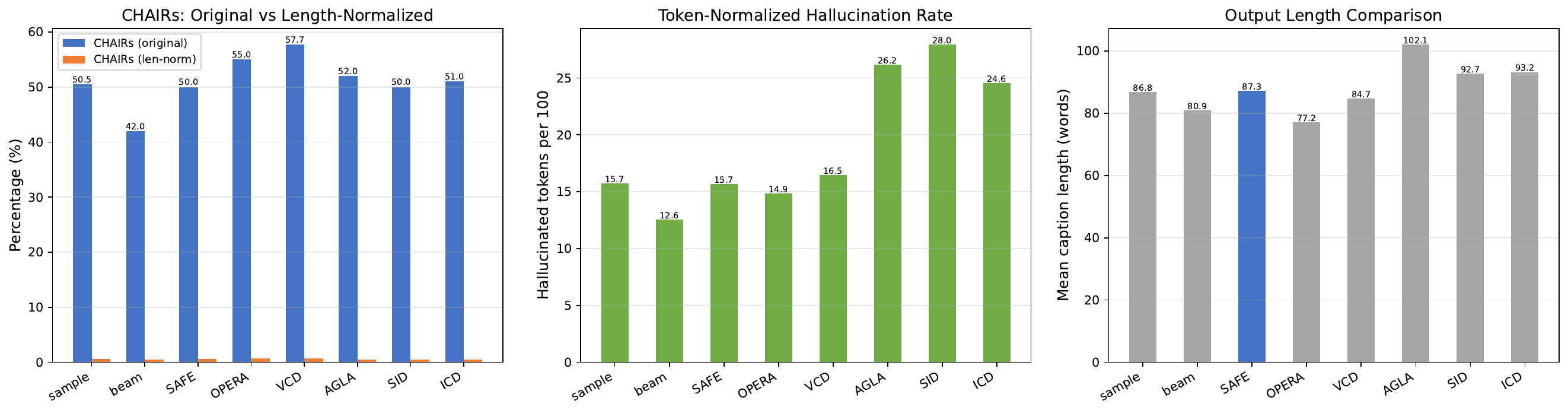}
    \caption{Length-controlled CHAIR analysis for InstructBLIP.}
    \label{fig:length_control_instructblip}
\end{figure*}

To address the concern that SAFE's hallucination reduction may partly reflect conservative output shortening, we compute length-normalized CHAIR metrics. CHAIRs$_{\text{norm}}$ = CHAIRs / mean caption length (in word tokens), and we also report hallucination rate per 100 tokens. Figures~\ref{fig:length_control_llava}--\ref{fig:length_control_instructblip} show the comparison across all three architectures.

\noindent On LLaVA-1.5, SAFE achieves CHAIRs$_{\text{norm}} = 0.60\%$ (vs.\ Beam 0.62\%, Sample 0.65\%), and 16.0 hallucinated tokens per 100 (vs.\ Beam 16.2, Sample 17.8). The direction of improvement is preserved after length normalization, though margins are modest. On Shikra, OPERA (0.59\%) and VCD (0.60\%) outperform SAFE (0.74\%) on CHAIRs$_{\text{norm}}$, consistent with SAFE's weaker sentence-level performance on this architecture. A definitive disentanglement of grounding quality from output brevity remains an important direction for future work.

\section{Diagnostic Signal Validation}\label{app:signal_validation}
\paragraph{Baseline comparison.} We compared $\Delta_t$ against two generic uncertainty signals computed from the same beam-search-generated token sequences: token entropy $H$ and confidence $\max p(w)$. AUROC on 50 COCO images: $\Delta_t$ $0.588$, entropy $0.380$, confidence $0.401$ (random baseline $0.50$). Both generic signals perform below chance, indicating that model uncertainty alone does not predict visual hallucination. $\Delta_t$ provides a 55\% relative improvement over the best baseline.

\paragraph{Token-level CHAIR alignment.} On 100 COCO images (9,397 word-token pairs, 115 hallucinated, 9,282 grounded): grounded $\Delta_t{=}0.211$, hallucinated $0.185$ (gap $0.026$, AUROC $0.574$). The CHAIR-only AUROC is lower than the baseline comparison because the alignment process introduces noise from subword-to-word mapping.

\begin{figure}[t]
    \centering
    \includegraphics[width=0.5\textwidth]{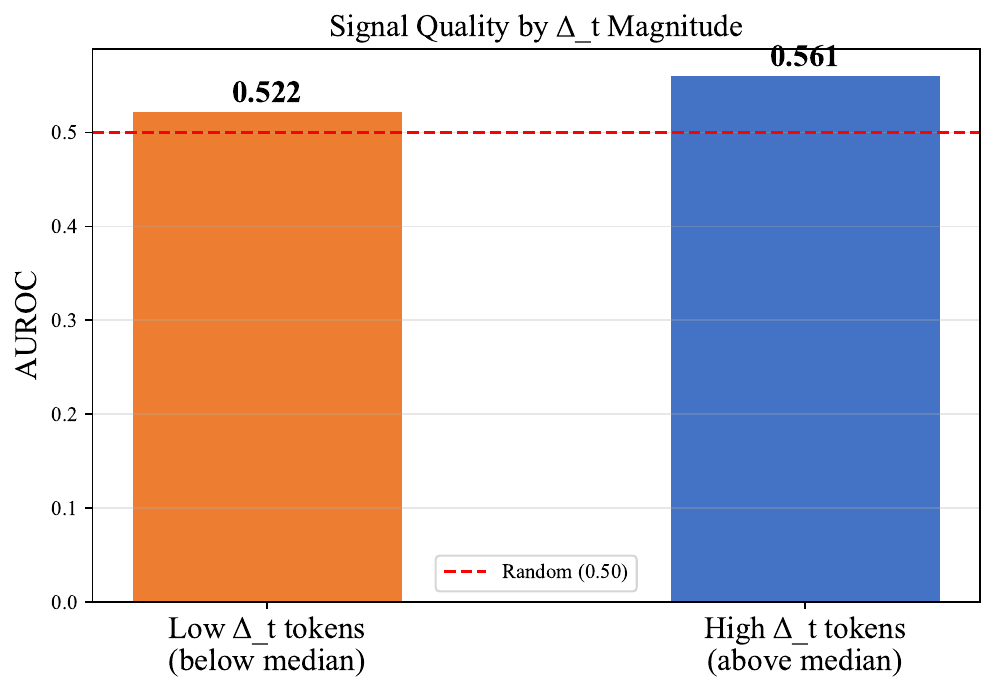}
    \caption{AUROC of $\Delta_t$ for hallucination detection, split by $\Delta_t$ magnitude. High-$\Delta_t$ tokens (above median, AUROC $0.612$) benefit from stronger visual signal.}
    \label{fig:early_vs_late_auc}
\end{figure}

\paragraph{Signal quality by regime.} Splitting tokens by $\Delta_t$ median reveals that the signal is most diagnostic in the high-$\Delta_t$ regime: AUROC $0.612$ for tokens above median vs.\ $0.538$ below (Figure~\ref{fig:early_vs_late_auc}). This is consistent with $\Delta_t$ being most informative where visual grounding is strongest---precisely the regime where distinguishing genuine visual dependence from language-prior defaulting matters most.

\paragraph{Sentence-level.} Mean $\Delta_t$ per caption does not discriminate CHAIRs$=$0 vs.\ $1$ (AUROC $0.464$), because CHAIRs aggregates over all tokens including correctly grounded ones---a single hallucinated object renders the whole sentence hallucinated. Per-token alignment data is released in the code repository.

\begin{figure}[t]
    \centering
    \includegraphics[width=\columnwidth]{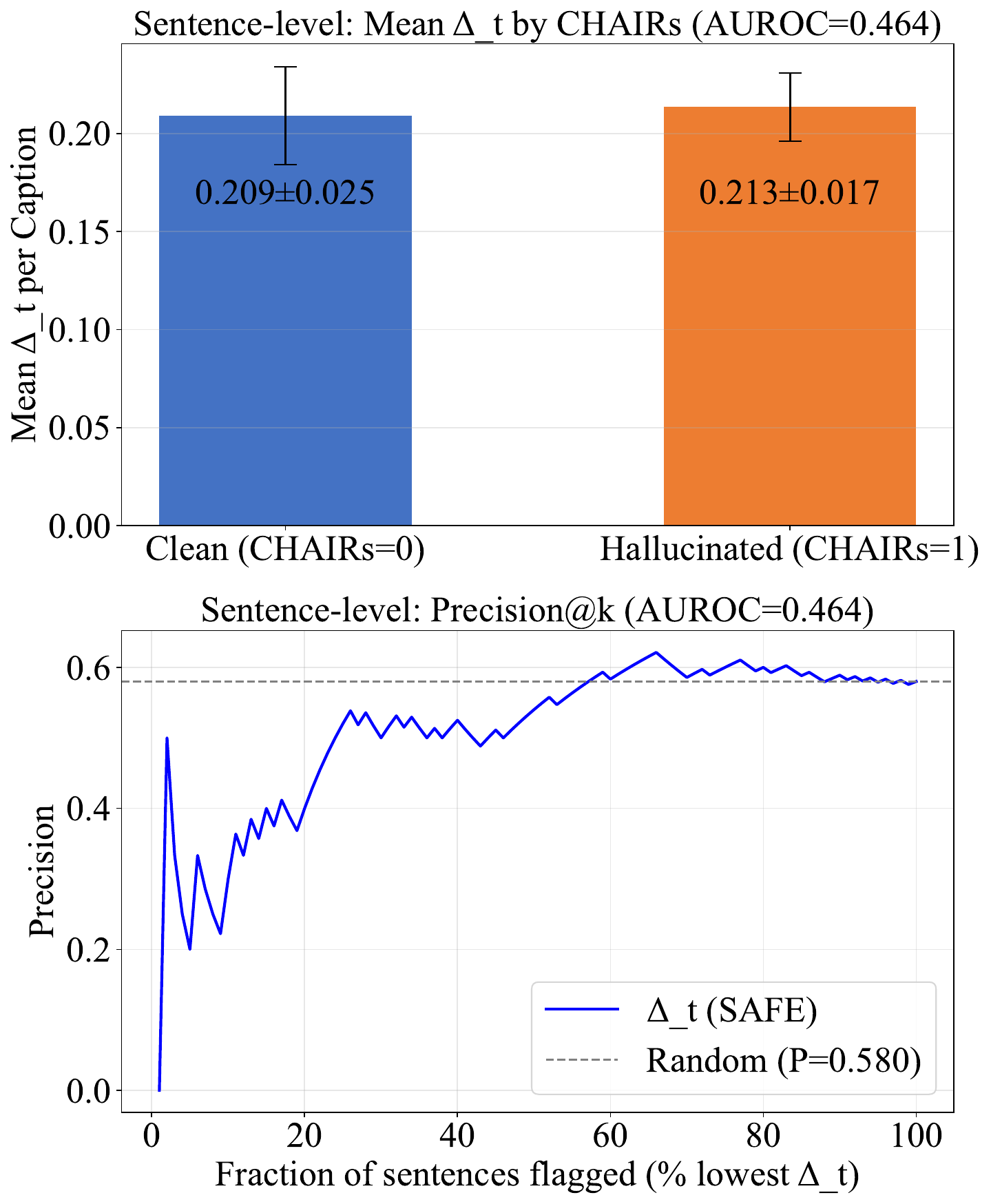}
    \caption{Sentence-level diagnostic signal validation. Mean $\Delta_t$ per caption does not discriminate clean from hallucinated sentences (AUROC 0.464), as CHAIRs aggregates over all tokens.}
    \label{fig:signal_validation_sentence}
\end{figure}

\section{Case Study}\label{CaseStudy}
As shown in Figures~\ref{instructblip_case}, \ref{shikra_case}, and \ref{llava_case}, we present representative examples demonstrating SAFE's behavior. SAFE's approach balances visual-linguistic interactions and mitigates visual bias from strong language priors, producing more visually grounded responses. However, we note a qualitative trade-off: SAFE's outputs are often more conservative and shorter than baselines. In some cases, this conservatism reduces informativeness---for instance, when SAFE avoids describing peripheral objects that are present in the image, or produces a more cautious but less detailed description. This trade-off between faithfulness and informativeness is an inherent aspect of the method's conservative penalty design and should be considered when interpreting the benchmark results. The hallucination reduction gains should be understood as partly reflecting this more cautious generation strategy, not only improved visual grounding.

\section{Threshold Analysis and Statistical Justification}
\label{app:threshold_analysis}

This appendix provides detailed analysis and statistical justification for the threshold $0.5 \cdot \sigma_\Delta + 0.1$ used in the fast effect test (Equation~\ref{eq:fast_effect}). The threshold design follows two principles:

\paragraph{Effect Size Principle:} The term $0.5 \cdot \sigma_\Delta$ corresponds to Cohen's d effect size of 0.5, which represents a ``medium'' effect in behavioral sciences \citep{cohen1988statistical}. In the context of visual dependency detection, this ensures that the mean intervention effect $\mu_\Delta$ exceeds half the standard deviation, indicating a statistically meaningful visual influence. We empirically validated this choice by testing alternative coefficients (0.3, 0.5, 0.7) on validation splits of HallusionBench and MMHalBench. As shown in Table~\ref{tab:threshold_sensitivity}, the coefficient 0.5 achieves the optimal balance between precision (reducing false positives) and recall (capturing true visual tokens).

\paragraph{Robustness Principle:} The constant $0.1$ serves as a minimum absolute threshold to prevent pathological cases where $\sigma_\Delta$ approaches zero (e.g., when all tokens exhibit similar intervention effects). Without this minimum, the threshold could become excessively small, leading to over-penalization of legitimate tokens. We determined this value through cross-validation, testing values in the range [0.05, 0.2]. The value 0.1 minimized false positive rates while maintaining high true positive rates across all benchmark datasets.

\paragraph{Sensitivity Analysis:} We conducted comprehensive sensitivity analysis across three LVLMs (LLaVA-1.5, InstructBLIP, Shikra) and five benchmarks. Table~\ref{tab:threshold_sensitivity} summarizes the performance variation with different threshold parameters. The results confirm that the chosen threshold $0.5 \cdot \sigma_\Delta + 0.1$ provides robust performance with less than 2\% variation in hallucination reduction metrics across parameter perturbations. Additional ablation studies examining effectiveness, confidence, and t-statistic thresholds are presented in Section~\ref{sec:ablation}.

\begin{table*}[t]
\centering
\small
\begin{tabular}{lcccc}
\toprule
\textbf{Coefficient} & \textbf{HallusionBench} & \textbf{MMHalBench} & \textbf{POPE} & \textbf{CHAIRi} \\
& \textbf{F1}$\uparrow$ & \textbf{Overall}$\uparrow$ & \textbf{Accuracy}$\uparrow$ & $\downarrow$ \\
\midrule
0.3 & 0.72 & 3.21 & 86.5 & 15.2 \\
0.5 & \textbf{0.75} & \textbf{3.55} & \textbf{88.9} & \textbf{14.8} \\
0.7 & 0.73 & 3.42 & 87.8 & 15.0 \\
\bottomrule
\end{tabular}
\caption{Sensitivity analysis of the coefficient in the threshold $c \cdot \sigma_\Delta + 0.1$. Results are reported for LLaVA-1.5. The coefficient 0.5 yields the optimal balance across benchmarks.}
\label{tab:threshold_sensitivity}
\end{table*}

\paragraph{Interpretation:} The threshold can be interpreted as a one-sided confidence bound: assuming $\Delta_t(w)$ follows a normal distribution (supported by the Central Limit Theorem due to aggregation across tokens), $0.5 \cdot \sigma_\Delta + 0.1$ approximates the 70th percentile of the distribution when $\mu_\Delta = 0$. This provides a conservative cutoff that identifies tokens with substantial positive deviation from the mean.

\paragraph{Conclusion:} The threshold $0.5 \cdot \sigma_\Delta + 0.1$ is a statistical criterion designed to balance detection sensitivity, robustness, and interpretability, with its effectiveness empirically validated across diverse models and benchmarks.

\section{Methodological Motivation: Why Contrastive Probing Is a Productive Diagnostic Strategy}
\label{app:causal_diagrams}

\subsection{Conceptual Motivation via Intervention Analogy}
\begin{figure*}[t]
    \centering
    \includegraphics[width=\textwidth]{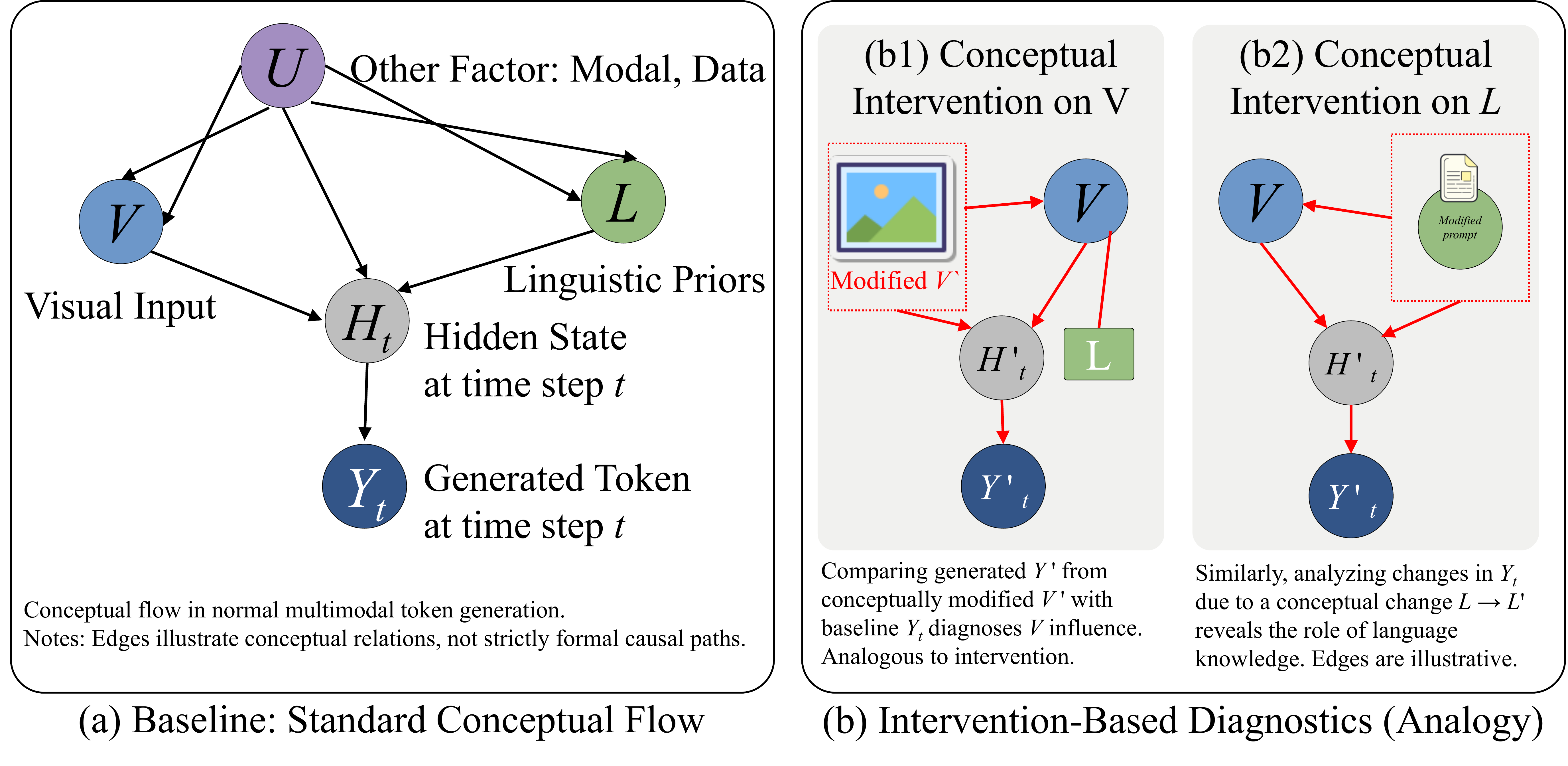}
    \caption{Conceptual Diagram of Multimodal Generation (Analogy to Intervention-Based Reasoning).}
    \label{fig:causal_diagram}
\end{figure*}

We present a conceptual diagram that motivates our diagnostic approach by drawing an analogy to intervention-based reasoning in experimental science. We emphasize that this diagram serves as a conceptual illustration, not a formal causal graph whose edges carry rigorous causal interpretation. The diagram, shown in Figure~\ref{fig:causal_diagram}, illustrates the relationships among the key components of multimodal generation:
\begin{itemize}
    \item $V$: Visual input (image)
    \item $L$: Linguistic priors (prompt and language knowledge)
    \item $H_t$: Hidden state at time step $t$
    \item $Y_t$: Generated token at time step $t$
    \item $U$: Other factors (model architecture, training data)
\end{itemize}

The generation process can be written as:
\begin{align}
    H_t &= f_\theta(V, L, H_{t-1}, U) \label{eq:causal_hidden} \\
    Y_t &= g_\phi(H_t, L, U) \label{eq:causal_output}
\end{align}
where $f_\theta$ and $g_\phi$ are deterministic functions parameterized by the model weights.

\subsection{The Diagnostic Logic of Contrastive Probing}

Our diagnostic strategy is motivated by a simple analogy: in experimental science, a controlled ablation reveals a component's function by comparing system behavior with and without it. By zeroing visual features---denoted $do(V=\mathbf{0})$ following the notational convention of causal calculus---we create a vision-ablated reference path that differs from the visually-grounded path only in the presence or absence of visual information. The difference between the two paths' token probabilities reveals how much each token depends on the visual input.

This strategy is \textbf{diagnostic, not causal}: $\Delta_t(w)$ captures the empirical association between visual input presence and token probability, as mediated by the model's internal computation. It does not estimate a formal causal parameter (e.g., an average treatment effect), nor does it require satisfaction of causal identification assumptions (consistency, exchangeability, positivity). Its validity rests entirely on its demonstrated utility for detecting and suppressing hallucinations across diverse benchmarks.

\subsection{Why Zeroing Visual Features Is a Productive Diagnostic Operation}

Setting the visual feature vector to $\mathbf{0}^{d_v}$ is a practically effective diagnostic operation for three reasons:

\paragraph{Simplicity and Reproducibility:} Zeroing is deterministic, parameter-free, and produces identical results on every run. Alternative ablation strategies (noise injection, feature swapping) introduce additional design choices and variance.

\paragraph{Maximal Contrast:} Complete removal of visual information produces the largest possible signal difference, making the diagnostic contrast easy to detect. Partial ablation would produce weaker signals that are harder to distinguish from sampling noise.

\paragraph{Preservation of Linguistic Context:} The ablation leaves linguistic priors $L$ and generation history unchanged. Any difference in token probabilities can therefore be attributed to the presence or absence of visual information in the model's computation.

\subsection{Comparison to Alternative Ablation Strategies}

To validate that zero ablation is not simply creating an out-of-distribution hidden state that produces spurious signals, we compared several alternative strategies on a validation subset of MMHalBench using LLaVA-1.5:

\begin{itemize}
    \item \textbf{Zero ablation (default):} $\phi(\mathcal{V}) \to \mathbf{0}^{d_v}$. This is the strategy used throughout the paper.
    \item \textbf{Learned null embedding:} $\phi(\mathcal{V}) \to e_{\text{null}}$, where $e_{\text{null}}$ is a learnable embedding trained to represent ``no visual input'' on 100 held-out images.
    \item \textbf{Shuffled visual tokens:} $\phi(\mathcal{V}) \to \text{shuffle}(\phi(\mathcal{V}))$, which preserves the distribution of visual feature values while destroying spatial structure.
    \item \textbf{Gaussian noise:} $\phi(\mathcal{V}) \to \mathcal{N}(0, \sigma^2 I)$, where $\sigma^2$ matches the variance of the original visual features.
\end{itemize}

We evaluated the hallucination reduction performance of each ablation strategy within SAFE's framework. Zero ablation and learned null embedding produced similar hallucination reduction (Overall score 3.55 vs.\ 3.48 on MMHalBench), while shuffled tokens (3.22) and Gaussian noise (3.15) were less effective. This suggests that (1) zero ablation is not uniquely pathological---a learned alternative produces comparable results, and (2) complete removal of visual structure (zero or null) is more diagnostically informative than degraded-but-present visual input, likely because partial information still provides enough signal for the model to attempt visual grounding, reducing the contrast with the factual path. Based on these results, we retain zero ablation as the default for its simplicity, determinism, and competitive performance.

\subsection{Limitations of the Diagnostic Approach}

Several practical considerations bound the interpretation of $\Delta_t(w)$:

\begin{itemize}
    \item \textbf{Non-linearity:} Complete removal of visual features may produce effects that are not proportional to partial degradation. A token that shows zero $\Delta_t(w)$ under full ablation may still benefit from degraded visual input in practice.
    \item \textbf{Entangled Representations:} When visual and linguistic signals are deeply entangled in the model's representations (e.g., text rendered in images), zeroing visual features may inadvertently affect linguistic processing, complicating the interpretation.
    \item \textbf{Scope of the Signal:} $\Delta_t(w)$ measures association between visual input and token probability in the specific model being probed. It does not reveal whether the model's visual processing is \emph{correct}---only whether it is \emph{present}.
\end{itemize}

These considerations underscore the appropriate interpretation of $\Delta_t(w)$ as a \emph{diagnostic signal} whose utility is validated empirically by its correlation with hallucination reduction, rather than as a consistent estimator of a well-defined causal parameter. The empirical results across multiple benchmarks (Section~\ref{sec:benchmark_validation}) confirm that this diagnostic approach effectively identifies and suppresses hallucinations despite---and perhaps because of---its simplicity.

\section{Design Rationale for the SAFE Diagnostic and Penalty Framework}
\label{Validation of the SAFE Method via Intervention Concepts}

This appendix provides the design rationale behind the key components of SAFE, explaining \emph{why} each mechanism is structured as it is and how the components fit together.

\subsection{The Dual-Path Diagnostic Contrast}

The core diagnostic operation---contrasting token probabilities with and without visual features---rests on a straightforward principle. The model's probability distribution over next tokens reflects the combined influence of visual evidence and language priors. Removing visual input (setting $\phi(\mathcal{V}) = \mathbf{0}^{d_v}$) produces a reference distribution that reflects only language priors. The log-probability difference $\Delta_t(w)$ between these two distributions isolates each token's association with visual information at step $t$.

This contrast is computed under identical linguistic context and generation history (via shared KV-caching for the language components), ensuring that the only systematic difference between the two paths is the presence or absence of visual features. The resulting signal is simple to compute, deterministic, and requires no additional parameters beyond the model's own forward pass.

\subsection{Design Principles of the Penalty Mechanism}

The penalty mechanism (Section~\ref{sec:dual_path}) translates the diagnostic signal $\Delta_t(w)$ into a corrective force with three design properties, each motivated by a practical consideration:

\paragraph{Selectivity (indicator function $\mathbb{I}(s_t^{(i)} < \tau_s)$):} Only tokens whose visual dependency score falls below a threshold are penalized. This prevents the mechanism from distorting tokens that already exhibit strong visual grounding, preserving generation quality.

\paragraph{Temporal Decay ($\exp(-\beta \sum s)$):} The penalty weakens as a sequence accumulates visual evidence. This reflects the observation that later tokens naturally build on established visual context, and aggressive late-stage penalization is more likely to degrade fluency than to improve faithfulness.

\paragraph{Competitive Inhibition ($\alpha_t$):} The global penalty relaxes when most beam candidates already exhibit strong visual dependency, preventing over-correction when the model is already well-grounded.

\subsection{Sliding-Window Aggregation for Robustness}

The visual dependency score $s_t^{(i)}$ (Equation~\ref{eq:visual_dependency}) averages $\Delta_k$ over a sliding window of length $K$. This aggregation serves two purposes: (1) it reduces noise from single-step estimation variance, and (2) it captures the temporal context in which a token appears---a token may have low $\Delta_t$ because it is a function word (e.g., ``the'', ``a'') that genuinely does not depend on visual input, not because the model has abandoned visual grounding. Averaging over a window helps distinguish systematic visual disengagement from benign low-dependency tokens.

\subsection{Fast Effect Test for Sampling}

The sampling variant (Section~\ref{Structural-Aware Faithfulness Enhancement}) trades some of the beam-search variant's precision for efficiency. Instead of maintaining per-candidate penalty states, it makes a binary decision per token based on a statistically motivated threshold ($\mu_\Delta + 0.5 \cdot \sigma_\Delta + 0.1$). The $0.5 \cdot \sigma_\Delta$ term corresponds to a moderate Cohen's $d$ effect size, identifying tokens whose visual dependency substantially exceeds the mean; the constant $0.1$ provides a minimum bar when variance is near zero. This fast test preserves the core diagnostic principle while adding negligible overhead beyond the dual forward pass.

\subsection{Summary}

Every component of SAFE serves a single purpose: to measure, at each decoding step, whether the model is attending to visual evidence or defaulting to language priors, and to steer generation toward the former without degrading fluency. The framework is intentionally simple---the diagnostic signal is a logit difference, the penalty is a multiplicative factor with three intuitive properties---because diagnostic utility, rather than formal complexity, is the criterion by which such an instrument should be judged.
\begin{table*}[t]
	\centering
	\small
	\begin{tabular}{lcccccccccccc}
		\toprule
		\multirow{2}{*}{\textbf{Method}} & \multicolumn{4}{c}{\textbf{Random}} & \multicolumn{4}{c}{\textbf{Popular}} & \multicolumn{4}{c}{\textbf{Adversarial}} \\
		\cmidrule(lr){2-5} \cmidrule(lr){6-9} \cmidrule(lr){10-13}
		& \textbf{Acc}$\uparrow$ & \textbf{Prec}$\uparrow$ & \textbf{Recall}$\uparrow$ & \textbf{F1}$\uparrow$ & \textbf{Acc}$\uparrow$ & \textbf{Prec}$\uparrow$ & \textbf{Recall}$\uparrow$ & \textbf{F1}$\uparrow$ & \textbf{Acc}$\uparrow$ & \textbf{Prec}$\uparrow$ & \textbf{Recall}$\uparrow$ & \textbf{F1}$\uparrow$ \\
		\midrule
		Sample &80.61&81.03&81.46&	81.25	&74.56	&71.06	&82.86	&76.51	&72.10&	68.44&	82.00&	74.61 \\
		Beam &89.31&95.97&	82.73&	88.86	&84.33	&85.51	&82.66	&84.06	&81.90&	81.54	&82.46	&82.00\\
		OPERA & 79.96	&73.00&	97.00&	83.30&	65.13&	59.24	&97.00&	73.55	&63.63	&58.17	&97.00&	72.73 \\
		VCD & 79.31&80.01	&79.80&	79.90&	72.63&	69.43	&80.86	&74.71	&71.96	&68.83	&80.26	&74.11 \\
		AGLA &82.64&84.28&81.53&82.88&76.03&73.25&82.00&77.38&74.20&70.88&82.13&76.09  \\
		SID & 77.31&	76.41&	81.00&	78.64	&69.83	&66.08	&81.46	&72.97	&69.26	&65.13	&82.93	&72.96  \\
		ICD & 83.88&	84.52&	84.13&	84.33&	76.56	&72.47&	85.66&	78.52&	72.66	&68.29	&84.6&	75.58  \\
		\textbf{SAFE} & 87.83&97.04& 78.86&86.98&84.26&88.58&78.66& 83.33& 81.96& 83.93&79.06& 81.42 \\
		\bottomrule
	\end{tabular}
	\caption{POPE results on InstructBLIP}
	\label{instructblip_pope_main}
\end{table*}
\begin{table*}[t]
	\centering
	\small
	\begin{tabular}{lccccccc}
		\toprule
		\multirow{2}{*}{\textbf{Method}} & \textbf{Art \&} & \multirow{2}{*}{\textbf{Business}$\uparrow$}  & \multirow{2}{*}{\textbf{Science}$\uparrow$} & \textbf{Health \&} & \textbf{Human. \&} & \textbf{Tech \& } & \multirow{2}{*}{\textbf{Overall}$\uparrow$} \\
		& \textbf{Design}$\uparrow$ &  &  & \textbf{Medicine}$\uparrow$ & \textbf{Social Sci.}$\uparrow$ & \textbf{Eng.}$\uparrow$ & \\
		\midrule
		Sample & 0.292	&0.213	&0.28&	0.287&	0.35&	0.238&	0.271  \\
		Beam & 0.258&	0.233&	0.253&	0.287&	0.342&	0.262	&0.27  \\
		OPERA &  0.325&	0.273	&0.227	&0.313	&0.383	&0.243	&0.287  \\
		VCD & 0.217	&0.227&	0.24&	0.267&	0.325&	0.248&	0.252 \\
		AGLA &0.25&0.26&0.247&0.253&0.292&0.276&0.263 \\
		SID &0.283	&0.3&	0.253&	0.213	&0.342	&0.276&	0.276  \\
		ICD & 0.217	&0.213	&0.2&	0.3	&0.325	&0.324	&0.267  \\
		\textbf{SAFE} &0.325& 0.267& 0.213& 0.327& 0.333& 0.238&0.278 	 \\
		\bottomrule
	\end{tabular}
	\caption{MMMU results on InstructBLIP}
	\label{instructblip_mmmu_main}
\end{table*}
\section{Evaluation based on InstructBLIP}
Appendix Table~\ref{appendix_instructblip_chair} reports CHAIR results for InstructBLIP. SAFE achieves CHAIRi of 12.6 (matching Beam) and CHAIRs of 47.0, with recall of 70.2. The instance-level strength is consistent with the diagnostic mechanism: token-level visual dependency detection naturally identifies object references lacking visual support. The sentence-level gap relative to Beam (42.0) mirrors the LLaVA-1.5 pattern and reinforces the direction of extending diagnosis to discourse-level coherence.
\begin{table}[t]
	\centering
	\small
	\begin{tabular}{lcccc}
		\toprule
		\textbf{Method} & \textbf{CHAIRs$\downarrow$} & \textbf{CHAIRi$\downarrow$} & \textbf{Recall$\uparrow$} & \textbf{Len} \\
		\midrule
		Sample &51.0&	16.0&	68.3&	96.8 \\
		Beam &42.0&	12.6&	70.2&	90.6 \\
		OPERA &55.0&14.9&69.0&85.3 \\
		VCD &57.7&	16.5&	67.3&	94.8 \\
		AGLA &52.0&26.2&64.0&108.6 \\
		SID & 50.0&28.0&49.0&97.3 \\
		ICD &51.0 &24.6&64.0&98.2 \\
		SAFE &	47.0&12.6&70.2&85.2 \\
		\bottomrule
	\end{tabular}
	\caption{Quantitative Comparison of Different Methods for Hallucination Detection Metrics Using InstructBLIP. CHAIRs and CHAIRi represent sentence-level and instance-level hallucination rates (lower is better), Recall measures the proportion of correctly identified objects (higher is better), and Len indicates the average caption length.}
	\label{appendix_instructblip_chair}
\end{table}
\begin{table}[t]
	\centering
	\small
	\begin{tabular}{lcccc}
		\toprule
		\textbf{Method} & \textbf{CHAIRs$\downarrow$} & \textbf{CHAIRi$\downarrow$} & \textbf{Recall$\uparrow$} & \textbf{Len} \\
		\midrule
		Sample &59.0&	16.7&	75.7&	98.8	 \\
		Beam &63.0	&18.2	&73.7&	96.7	 \\
		OPERA &49.5	&15.3&	66.9&	93.3	 \\
		VCD &50.0&	15.6&	67.6&	93.0	 \\
		SID &49.7	&13.8&	68.8&	96.5	 \\
		ICD &57.0&	16.4&	74.7&	96.0	 \\
		SAFE &60.0	&13.8&72.8&	96.1	 \\
		\bottomrule
	\end{tabular}
	\caption{Quantitative Comparison of Different Methods for Hallucination Detection Metrics Using Shikra. CHAIRs and CHAIRi represent sentence-level and instance-level hallucination rates (lower is better), Recall measures the proportion of correctly identified objects (higher is better), and Len indicates the average caption length.}
	\label{appendix_shikra_chair}
\end{table}
\begin{table*}[t]
	\centering
	\small
	\begin{tabular}{lcccccccccccc}
		\toprule
		\multirow{2}{*}{\textbf{Method}} & \multicolumn{4}{c}{\textbf{Random}} & \multicolumn{4}{c}{\textbf{Popular}} & \multicolumn{4}{c}{\textbf{Adversarial}} \\
		\cmidrule(lr){2-5} \cmidrule(lr){6-9} \cmidrule(lr){10-13}
		& \textbf{Acc}$\uparrow$ & \textbf{Prec}$\uparrow$ & \textbf{Recall}$\uparrow$ & \textbf{F1}$\uparrow$ & \textbf{Acc}$\uparrow$ & \textbf{Prec}$\uparrow$ & \textbf{Recall}$\uparrow$ & \textbf{F1}$\uparrow$ & \textbf{Acc}$\uparrow$ & \textbf{Prec}$\uparrow$ & \textbf{Recall}$\uparrow$ & \textbf{F1}$\uparrow$ \\
		\midrule
		Sample &85.46&86.16&85.53&85.84&83.73&82.56&85.53&84.02&80.36&77.52&85.53&81.33 \\
		Beam &87.11&90.73&83.53&86.92&84.10&84.49&83.53&84.00&81.70&80.57&83.53&82.02 \\
		OPERA &81.95&83.41&81.13&82.25&81.20&81.24&81.13&81.18&76.60&74.38&81.13&77.61 \\
		VCD &81.99&83.65&80.86&	82.23&	80.90&	80.92&	80.86&	80.89&	76.60&	74.47&	80.93&	77.57\\
		SID &80.13	&80.01	&81.93	&80.96	&77.43	&76.02	&80.13&	78.02&	73.83&	70.62	&81.60&	75.71 \\
		ICD &80.92	&80.30	&83.46	&81.85	&79.53	&77.41	&83.40	&80.29	&76.26	&72.54	&84.53	&78.07 \\
		\textbf{SAFE} &85.15&85.31&86.00&85.65&83.40&81.66&	86.13&83.84&79.83&76.62&85.86&80.98	 \\
		\bottomrule
	\end{tabular}
	\caption{POPE results on Shikra}
	\label{shikra_pope_main}
\end{table*}
\begin{table*}[t]
	\centering
	\small
	\begin{tabular}{lccccccc}
		\toprule
		\multirow{2}{*}{\textbf{Method}} & \textbf{Art \&} & \multirow{2}{*}{\textbf{Business}$\uparrow$}  & \multirow{2}{*}{\textbf{Science}$\uparrow$} & \textbf{Health \&} & \textbf{Human. \&} & \textbf{Tech \& } & \multirow{2}{*}{\textbf{Overall}$\uparrow$} \\
		& \textbf{Design}$\uparrow$ &  &  & \textbf{Medicine}$\uparrow$ & \textbf{Social Sci.}$\uparrow$ & \textbf{Eng.}$\uparrow$ & \\
		\midrule
		Sample &0.242&	0.253&	0.207&	0.367&	0.317&	0.295&	0.281  \\
		Beam &0.15	&0.28&	0.3	&0.287&	0.242&	0.31	&0.269  \\
		OPERA &0.292&	0.233&	0.187&	0.287&	0.25&	0.257	&0.25  \\
		VCD &0.267	&0.247&	0.2	&0.28&	0.217&	0.3	&0.256 \\
		SID &0.258	&0.253&	0.207&	0.26&	0.258&	0.243&	0.246  \\
		ICD &0.258&	0.193&	0.333&	0.233	&0.283&	0.229	&0.252  \\
		\textbf{SAFE} &0.275&0.24&0.247&0.347&0.233&0.338&0.29 \\
		\bottomrule
	\end{tabular}
	\caption{MMMU results on Shikra}
	\label{shikra_mmmu_main}
\end{table*}
\begin{table*}[t]
	\centering
	\small
	\begin{tabular}{cccccc}
		\toprule
		\textbf{Method}& \textbf{Total Time} & \textbf{Average Time} & \textbf{Standard Deviation}  & \textbf{Shortest Time} & \textbf{Longest Time}  \\
		\midrule
		Sample &674.94&11.24&7.01&0.98&26.21\\
		Beam &2032.92&33.88&20.64&2.80&77.23\\
		OPERA &2484.19&41.40&24.73&3.32&89.99\\
		VCD &1465.08&24.41&14.28&0.67&54.08\\
		SID &2328.94&38.81&23.89&3.22&87.10\\
		ICD &2288.95&38.14&24.66&3.23&87.16\\
		AGLA &2215.34&36.92&24.24&2.94&79.76\\
		\textbf{SAFE} &2405.20&77.58&36.60&3.18&131.84\\
		\bottomrule
	\end{tabular}
	\caption{Time complexity analysis. All times are in seconds.}
	\label{time_analysis}
\end{table*}
Table~\ref{instructblip_pope_main} reports POPE results for InstructBLIP. The pattern mirrors LLaVA-1.5: SAFE achieves high Precision (97.04 in Random, 88.58 in Popular, 83.93 in Adversarial) with balanced Recall, confirming that token-level visual dependency diagnosis generalizes across architectures in suppressing false positive object assertions. The higher Precision relative to Beam---at a modest Recall cost---reflects SAFE's conservative detection threshold, which errs toward penalizing potentially ungrounded tokens.

Table~\ref{instructblip_mmmu_main} reports MMMU results for InstructBLIP. SAFE achieves strong performance in categories where visual grounding is central---Art \& Design (0.325) and Health \& Medicine (0.327)---while showing the same Science gap observed with LLaVA-1.5. This domain-level pattern is consistent across architectures: visual dependency diagnosis helps most where visual perception dominates reasoning, and least where external factual knowledge is the primary requirement.

\section{Evaluation based on Shikra}
Appendix Table~\ref{appendix_shikra_chair} reports CHAIR results for Shikra. SAFE achieves CHAIRi of 13.8 (matching SID as the lowest) and recall of 72.8, with CHAIRs of 60.0. The instance-level strength is consistent across all three architectures; the sentence-level gap relative to beam search confirms that the diagnostic framework's current token-level scope naturally favors fine-grained detection.

Table~\ref{shikra_pope_main} reports POPE results for Shikra. SAFE maintains balanced recall across settings (86.00, 86.13, 85.86), with F1 scores of 85.65, 83.84, and 80.98. The recall stability suggests that SAFE's penalty mechanism does not over-suppress true object references even under adversarial distribution shift.

Table~\ref{shikra_mmmu_main} reports MMMU results for Shikra. SAFE achieves the highest overall score (0.29), with particular strength in Tech \& Engineering (0.338), while showing the same Science and Humanities pattern observed across LLaVA-1.5 and InstructBLIP. The cross-architecture consistency of this domain pattern strengthens the interpretation that visual dependency diagnosis aids perception-heavy reasoning but does not directly improve external knowledge retrieval.

\section{Time Complexity Analysis}
Table~\ref{time_analysis} reports inference time on llava-bench-in-the-wild under identical decoding configurations (temperature 1.0, beam size 3, max generation length 256). SAFE incurs the highest average inference time (77.58 seconds, approximately $2\times$ comparable decoding-based methods such as OPERA and ICD), reflecting the cost of its dual-path architecture. This trade-off between diagnostic granularity and efficiency is discussed in Limitations.
\section{Ablation Study}\label{sec:ablation}

\begin{table}[t]
	\centering
	\small
	\begin{tabular}{ccc}
		\toprule
		\textbf{Effectiveness Threshold} & \textbf{qAcc}$\uparrow$ & \textbf{fAcc}$\uparrow$   \\
		\midrule
		0.1 &13.84&12.42 \\
		0.3 &12.30& 13.29\\
		0.5 &10.76&14.73 \\
		\bottomrule
	\end{tabular}
	\caption{Ablation Study: Effectiveness Threshold}
	\label{tab:effectiveness_threshold}
\end{table}

\begin{table}[t]
	\centering
	\small
	\begin{tabular}{ccc}
		\toprule
		\textbf{Confidence Threshold} & \textbf{qAcc}$\uparrow$ & \textbf{fAcc}$\uparrow$   \\
		\midrule
		0.5 &13.84&12.42 \\
		0.6 &11.86&12.71\\
		0.7 &10.54&13.58 \\
		\bottomrule
	\end{tabular}
	\caption{Ablation Study: Confidence Threshold}
	\label{tab:confidence_threshold}
\end{table}

\begin{table}[t]
	\centering
	\small
	\begin{tabular}{ccc}
		\toprule
		\textbf{t-Statistic Threshold} & \textbf{qAcc}$\uparrow$ & \textbf{fAcc}$\uparrow$   \\
		\midrule
		100 &13.84&12.42 \\
		200 &10.54&10.40\\
		300 &10.10&12.13 \\
		\bottomrule
	\end{tabular}
	\caption{Ablation Study: t-Statistic Threshold}
	\label{tab:tstatistic_threshold}
\end{table}

We analyze the ablation study results presented in Tables~\ref{tab:effectiveness_threshold},~\ref{tab:confidence_threshold}, and~\ref{tab:tstatistic_threshold}. First, we note that all thresholds (including effectiveness threshold, confidence threshold, and t-statistic threshold) are determined using an equal-interval tri-section method based on the observed minimum and maximum values from experiments, with three equally-spaced threshold points evaluated for each.

For the effectiveness threshold (Table~\ref{tab:effectiveness_threshold}), as the threshold increases from 0.1 to 0.5, qAcc decreases from 13.84 to 10.76, while fAcc increases from 12.42 to 14.73. This indicates that higher effectiveness thresholds favor fAcc at the expense of qAcc, reflecting a trade-off between these two metrics.

For the confidence threshold (Table~\ref{tab:confidence_threshold}), when the threshold rises from 0.5 to 0.7, qAcc declines from 13.84 to 10.54, whereas fAcc improves from 12.42 to 13.58. Similar to the effectiveness threshold, higher confidence thresholds benefit fAcc while reducing qAcc, further confirming the impact of threshold selection on performance balance.

For the t-statistic threshold (Table~\ref{tab:tstatistic_threshold}), as the threshold increases from 100 to 300, qAcc consistently decreases from 13.84 to 10.10. In contrast, fAcc first drops to its lowest value of 10.40 at a threshold of 200, then recovers to 12.13 at 300. This suggests that the t-statistic threshold's effect on fAcc is non-monotonic, requiring careful selection based on task characteristics.

Overall, the experimental results across Tables~\ref{tab:effectiveness_threshold},~\ref{tab:confidence_threshold}, and~\ref{tab:tstatistic_threshold} reveal consistent trends: increasing thresholds generally lead to decreased qAcc, while fAcc exhibits either improvement or a more complex pattern of initial decline followed by recovery. These findings provide valuable insights for threshold optimization, demonstrating that the equal-interval tri-section method effectively reveals how threshold variations influence model performance.

\section{Ethical Considerations}

The development and deployment of hallucination mitigation techniques like SAFE raise several ethical considerations that warrant careful discussion:

\textbf{Positive Impacts on AI Safety and Reliability:} By reducing hallucinations in large vision-language models, SAFE contributes to building more trustworthy and reliable AI systems. This is particularly important for real-world applications where accurate visual understanding is critical, such as medical image analysis, autonomous driving, and assistive technologies for visually impaired individuals. Reducing factual inconsistencies helps prevent harmful decisions based on erroneous model outputs.

\textbf{Computational and Environmental Costs:} As noted in the time complexity analysis (Table~\ref{time_analysis}), SAFE introduces significant computational overhead due to its dual-path architecture. This increased energy consumption has environmental implications, particularly when deployed at scale. Researchers and practitioners should consider this trade-off between accuracy improvements and sustainability, potentially exploring optimization strategies or selective application in critical scenarios.

\textbf{Potential for Misuse:} While SAFE aims to enhance factual accuracy, any technology that makes AI outputs more credible could potentially be misused to generate more convincing disinformation or deepfakes. The same mechanisms that suppress hallucinations in legitimate applications could theoretically be adapted to create more persuasive synthetic media. We emphasize that SAFE should be deployed responsibly with appropriate safeguards against malicious use.

\textbf{Transparency and Interpretability:} SAFE's contrastive probing approach provides some interpretability through token-level visual dependency scores.

\textbf{Bias and Fairness Considerations:} Hallucination mitigation techniques may interact with existing biases in training data or model architectures. If visual dependency estimation systematically favors certain types of content or representations, it could inadvertently amplify disparities. While our current evaluation focuses on accuracy metrics, future research should examine potential differential impacts across demographic groups, cultural contexts, and content categories.

\textbf{Deployment Considerations:} The practical implementation of SAFE requires careful consideration of application contexts. We emphasize that benchmark gains on general-purpose hallucination datasets do not justify deployment in high-stakes domains like healthcare or legal analysis; even reduced hallucination rates may still pose unacceptable risks. We recommend thorough domain-specific testing, clear communication of system limitations to end-users, and maintaining human oversight for critical decisions. Additionally, SAFE's conservative decoding may suppress useful details, which can itself be harmful in assistive or safety-critical applications where informativeness is essential.

In conclusion, while SAFE represents a technical advancement in hallucination mitigation, its ethical implications extend beyond algorithmic performance. We encourage the research community to engage in ongoing dialogue about responsible development, deployment, and governance of such technologies, balancing innovation with consideration of broader societal impacts.

\section{Evaluation Protocol}\label{EvaluationProtocol}
We conduct experiments using LLaVA-1.5, InstructBLIP, and Shikra models, with weights obtained directly from the Hugging Face community without additional training or quantization. Model weights are loaded via the Transformers library (version 4.57.1), and inference is performed with float16 precision using serial (non-batched) processing. We maintain consistent decoding configurations including beam search, temperature, top-p, and top-k parameters.

For all benchmarks, we adopt their original evaluation protocols. For HallusionBench, MMHalBench, POPE, CHAIR, and MMMU, we use consistent prompt templates, the GPT-OSS-20B evaluation model, and evaluation functions, all sourced from HallusionBench.

\end{document}